\documentclass[journal]{IEEEtran}
\usepackage{amsmath,amsfonts}
\usepackage{url}
\usepackage[hidelinks,hypertexnames=false]{hyperref}
\usepackage{graphicx}
\usepackage{cite}
\usepackage{booktabs}
\usepackage{multirow}
\usepackage{makecell}
\usepackage{mathrsfs}
\usepackage{ulem}
\usepackage{color}
\usepackage{float}
\usepackage{subcaption}
\usepackage{caption}
\usepackage[switch]{lineno}
\usepackage{ulem}
\usepackage{anyfontsize}
\usepackage[table,xcdraw]{xcolor}
\usepackage{scalerel}
\usepackage{tikz}
\usetikzlibrary{svg.path}

\definecolor{orcidlogocol}{HTML}{A6CE39}
\tikzset{
  orcidlogo/.pic={
    \fill[orcidlogocol] svg{M256,128c0,70.7-57.3,128-128,128C57.3,256,0,198.7,0,128C0,57.3,57.3,0,128,0C198.7,0,256,57.3,256,128z};
    \fill[white] svg{M86.3,186.2H70.9V79.1h15.4v48.4V186.2z}
                 svg{M108.9,79.1h41.6c39.6,0,57,28.3,57,53.6c0,27.5-21.5,53.6-56.8,53.6h-41.8V79.1z M124.3,172.4h24.5c34.9,0,42.9-26.5,42.9-39.7c0-21.5-13.7-39.7-43.7-39.7h-23.7V172.4z}
                 svg{M88.7,56.8c0,5.5-4.5,10.1-10.1,10.1c-5.6,0-10.1-4.6-10.1-10.1c0-5.6,4.5-10.1,10.1-10.1C84.2,46.7,88.7,51.3,88.7,56.8z};
  }
}
\newcommand\orcidicon[1]{\href{https://orcid.org/#1}{\mbox{\scalerel*{
\begin{tikzpicture}[yscale=-1,transform shape]
\pic{orcidlogo};
\end{tikzpicture}
}{|}}}}
\newcolumntype{L}[1]{>{\raggedright\arraybackslash}p{#1}}
\newcolumntype{C}[1]{>{\centering\arraybackslash}p{#1}}
\newcolumntype{R}[1]{>{\raggedleft\arraybackslash}p{#1}}
\definecolor{TYS}{rgb}{0.6, 0.8, 0.2}
\definecolor{DOcolor}{rgb}{1,0.45,0.0}
\definecolor{NAVYcolor}{rgb}{0.05,0,0.5}

\newcommand{\settablefont}{\fontsize{6.5}
{10.0}\selectfont}

\newcommand{\eg}{\textit{e.g.}}
\newcommand{{\etal}}{\textit{et al.}}
\newcommand{\fixedtilde}[3]{%
  \begin{aligned}
    \rlap{$\tilde{\phantom{#1}}$}#1_{#2}^{#3}
  \end{aligned}
}
\newcommand{\fixedhat}[3]{%
  \begin{aligned}
    \rlap{$\hat{\phantom{#1}}$}#1_{#2}^{#3}
  \end{aligned}
}
\newcommand{\sra}{\raisebox{0.15ex}{\scalebox{0.7}{$\rightarrow$}}}
\begin{document}
\title{
Two-Stream Interactive Joint Learning of Scene Parsing and Geometric Vision Tasks
}
\normalem
\author{
Guanfeng Tang$^{\orcidicon{0009-0002-5918-4775}\,}$,
Hongbo~Zhao$^{\orcidicon{0009-0008-2198-2484}\,}$,
Ziwei~Long$^{\orcidicon{0009-0007-1520-5526}\,}$, 
Jiayao~Li$^{\orcidicon{0009-0006-5488-6662}\,}$, 
Bohong~Xiao$^{\orcidicon{0009-0008-6153-2605}\,}$,
Wei Ye$^{\orcidicon{0000-0002-3784-7788}}$,
Hanli~Wang$^{\orcidicon{0000-0002-9999-4871}\,}$,~\IEEEmembership{Senior Member,~IEEE},
Rui Fan$^{\orcidicon{0000-0003-2593-6596}\,}$,~\IEEEmembership{Senior Member,~IEEE}
    \vspace{-2.0em}
    \thanks{\textit{Corresponding author: Rui Fan}.}
    \thanks{Guanfeng Tang, Ziwei Long, and Wei Ye are with the College of Electronic and Information Engineering, Tongji University, Shanghai 201804, China (e-mail: \{2251778, zwlong, yew\}@tongji.edu.cn).}
    \thanks{Hongbo Zhao is with the Shanghai Research Institute for Intelligent Autonomous Systems, Tongji University, Shanghai 201210, China (e-mail: hongbozhao@tongji.edu.cn).}
    \thanks{Jiayao Li is with the College of Computer Science and Technology, Tongji University, Shanghai 201804, China (e-mail: lijiayao2351405@tongji.edu.cn).}
    \thanks{Bohong Xiao is with the Department of Vehicle Control System and Software Development, NIO 201804, Shanghai, China. He is also with the Department of Automotive Engineering, Jilin University, Jilin 130025, China (e-mail: bohong.xiao@nio.com).}
    \thanks{Hanli Wang is with the College of Electronic and Information Engineering, the School of Computer Science and Technology, and the Key Laboratory of Embedded System and Service Computing (Ministry of Education), Tongji University, Shanghai 201804, China (e-mail: hanliwang@tongji.edu.cn).}
    \thanks{
    Rui Fan is with the College of Electronic and Information Engineering, Shanghai Institute of Intelligent Science and Technology, Shanghai Research Institute for Intelligent Autonomous Systems, Shanghai Key Laboratory of Intelligent Autonomous Systems, State Key Laboratory of Autonomous Intelligent Unmanned Systems, and Frontiers Science Center for Intelligent Autonomous Systems (Ministry of Education), Tongji University, Shanghai 201804, China (e-mail: rui.fan@ieee.org).
}
}

\maketitle

\begin{abstract}
Inspired by the human visual system, which operates on two parallel yet interactive streams for contextual and spatial understanding, this article presents Two Interactive Streams (TwInS), a novel bio-inspired joint learning framework capable of simultaneously performing scene parsing and geometric vision tasks. 
TwInS adopts a unified, general-purpose architecture in which multi-level contextual features from the scene parsing stream are infused into the geometric vision stream to guide its iterative refinement. 
In the reverse direction, decoded geometric features are projected into the contextual feature space for selective heterogeneous feature fusion via a novel cross-task adapter, which leverages rich cross-view geometric cues to enhance scene parsing. 
To eliminate the dependence on costly human-annotated correspondence ground truth, TwInS is further equipped with a tailored semi-supervised training strategy, which unleashes the potential of large-scale multi-view data and enables continuous self-evolution without requiring ground-truth correspondences. 
Extensive experiments conducted on three public datasets validate the effectiveness of TwInS's core components and demonstrate its superior performance over existing state-of-the-art approaches. The source code will be made publicly available upon publication.
\end{abstract}


\begin{figure*}[!t]
    \centering
    \includegraphics[width=0.99\textwidth]{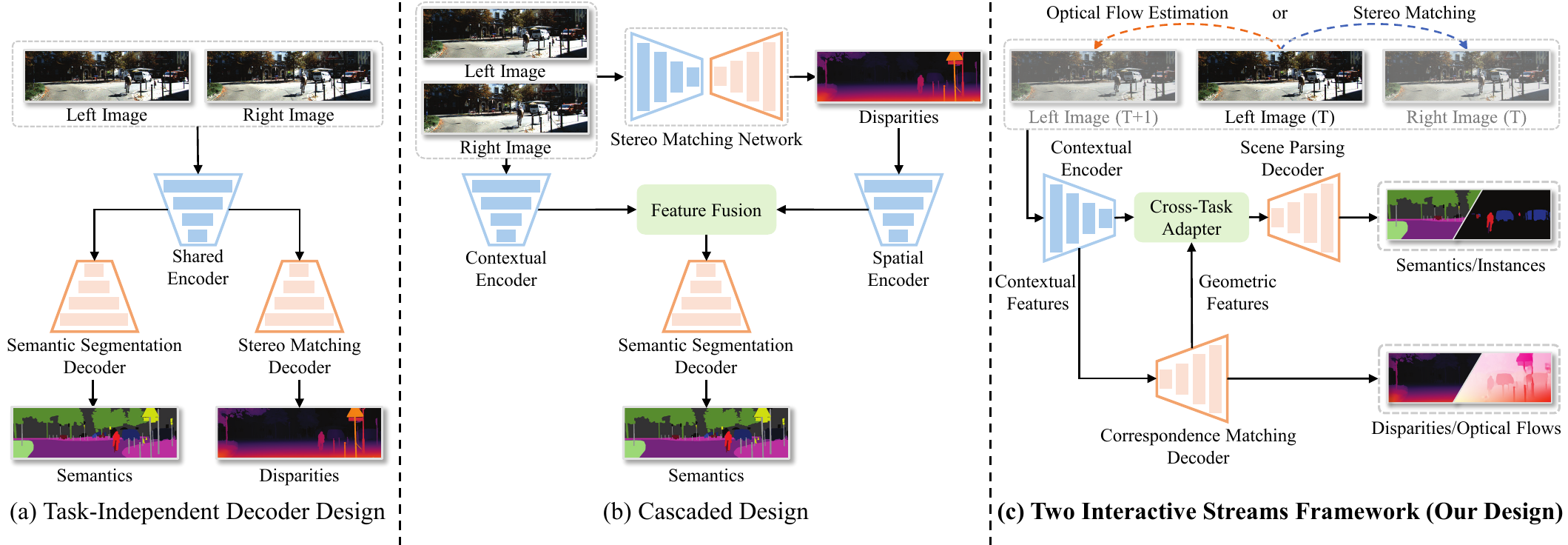}
    \caption{
    Comparison of joint learning frameworks: (a) traditional task-independent decoder design, where a shared encoder extracts visual features that are independently processed by task-specific decoders; (b) cascaded joint learning design, where disparities estimated by an iterative stereo matching network are fed into a feature-fusion semantic segmentation network via an additional spatial encoder; (c) the proposed general-purpose two-stream interactive framework, which enables bidirectional interaction by leveraging task-specific features to jointly optimize scene parsing and geometric vision tasks.
    }
    \vspace{-1.5em}
    \label{fig.framework}
\end{figure*}

\section{Introduction}
\label{sec.intro}
\IEEEPARstart{T}{he} \textit{two-streams hypothesis} posits that the human visual system consists of two functionally distinct pathways~\cite{twostream1}. The \textit{ventral stream} is responsible for scene parsing, whereas the \textit{dorsal stream}  perceives the 3D geometry of the environment. These two streams are closely interconnected, enabling humans to achieve comprehensive scene understanding. Inspired by this hypothesis, a growing body of studies~\cite{ticoss,sgroadseg,s3mnet,sgroadseg+,dsnet,SGDepth} in autonomous driving have developed humanoid visual perception systems, capable of jointly performing scene parsing tasks (\eg, semantic or instance segmentation) and geometric vision tasks (\eg, stereo matching or optical flow estimation) within a unified joint learning framework.

Recent studies~\cite{ticoss,sgroadseg} have specifically focused on the joint learning of semantic segmentation and stereo matching to emulate human-like visual perception. As illustrated in Fig.~\ref{fig.framework}(a), a representative class of frameworks adopts a single shared encoder to extract general-purpose visual features from RGB images~\cite{mtl_review1_tpami}. These features are subsequently fed into two task-specific decoders for independent predictions. However, most existing stereo matching methods capture cross-view geometric information primarily through cost volume aggregation or iterative refinement at the decoding stage, while the encoder remains limited to learning view-specific features. As a result, this task-independent decoder design impedes the effective integration of geometric cues into the semantic segmentation task, thereby limiting the potential performance gains achievable through deeper cross-task interaction.

With the rapid advancement of consumer-grade RGB-D sensors and multi-modal deep learning techniques~\cite{rgbd_sensors_tip,guo2024lix}, feature-fusion networks have achieved remarkable success in semantic segmentation by employing duplex encoders to extract complementary heterogeneous features from both RGB and depth images, which are then fused to achieve a more comprehensive understanding of the environment \cite{roadformer}. Recently, we took a step further by designing a cascaded, tightly coupled joint learning framework that integrates disparities estimated by an iterative stereo matching network into a feature-fusion semantic segmentation network through an independent spatial encoder~\cite{s3mnet,ticoss}, as illustrated in Fig. \ref{fig.framework}(b).
While this design inherently couples the two tasks and delivers notable improvements over previous joint learning methods, it still suffers from several critical limitations. First, within this cascaded framework, the informative contextual features learned for semantic segmentation do not feed back to improve stereo matching, leading to imbalanced task optimization. Second, the framework treats the estimated disparities merely as additional inputs, processed by an additional encoder. This design not only overlooks the rich decoded features in stereo matching networks, which aggregate cross-view geometric information through iterative refinement, but also introduces significant computational cost. Third, this framework remains confined to the joint learning of stereo matching and semantic segmentation, lacking the flexibility to extend to other scene parsing tasks (\eg, instance segmentation) and geometric vision tasks (\eg, optical flow estimation). This limitation reduces its versatility and practical applicability, particularly in scenarios where stereo cameras are unavailable or instance-level predictions are required.

To overcome the limitations of the aforementioned two frameworks, this study proposes a more general-purpose joint learning framework that is not restricted to stereo matching and semantic segmentation. As illustrated in Fig.~\ref{fig.framework}(c), the proposed framework consists of two bidirectionally interactive streams: a scene parsing stream based on Mask2Former~\cite{mask2former} for semantic or instance segmentation, and a geometric vision stream that leverages iterative refinement strategy for stereo matching or optical flow estimation.
On the one hand, multi-level contextual features learned for scene parsing tasks are selectively infused into the geometric vision stream to guide its iterative refinement process. 
On the other hand, decoded geometric features utilized for geometric vision tasks are fed back into the scene parsing stream to provide cross-view geometric cues. Nevertheless, a significant domain gap exists between the decoded geometric features and the contextual features, stemming from the distinct optimization objectives of the two task types~\cite{vitas}. Previous joint learning frameworks~\cite{s3mnet,ticoss,sgroadseg} that fuse heterogeneous features extracted by duplex encoders are thus incompatible with the proposed framework. To address this challenge, this study introduces a cross-task adapter~(CTA), which first projects geometric features into the contextual feature space and then selectively fuses such heterogeneous features, resulting in enriched features that incorporate both contextual and geometric cues.

As for the training strategy, most existing joint learning frameworks~\cite{semstereo_chen_aaai2025,s3mnet,ticoss} rely on fully supervised learning, necessitating both disparity and segmentation annotations. This dependence imposes significant constraints on dataset availability, making it impractical to train such frameworks on large-scale driving scene parsing datasets like Cityscapes~\cite{cityscapes}, which lack accurate disparity ground truth. Although recent joint learning studies~\cite{sgroadseg,sgroadseg+} have resorted to unsupervised stereo matching to tackle this problem, the commonly used photometric consistency loss provides only weak supervisory signals compared to the explicit pixel-wise constraints enforced in supervised approaches. This limitation becomes even more pronounced when extending to optical flow estimation, a more general 2D correspondence matching task, where the lack of strong supervision can significantly constrain the upper bound of performance. Such inaccurate geometric information may further compromise the performance of scene parsing tasks within a unified joint learning framework~\cite{roadformer}. To mitigate this issue, this study introduces a novel semi-supervised training strategy for geometric vision tasks, which can be seamlessly integrated into the proposed two-stream joint learning framework. 
Specifically, a teacher model pre-trained in a supervised manner on labeled synthetic data is employed to generate sparse yet high-quality labels for unlabeled real-world datasets. This is achieved by estimating the aleatoric uncertainty of the predicted disparities or optical flows and filtering out predictions with low confidence according to a quantile threshold derived from the uncertainty distribution. A student model with an identical architecture is then trained in a semi-supervised fashion using these pseudo labels, enabling the effective incorporation of large-scale multi-view data to improve the performance of both scene parsing and geometric vision tasks.
The above contributions collectively facilitate the development of \textbf{Tw}o \textbf{In}teractive \textbf{S}treams (\textbf{TwInS}), a novel two-stream interactive joint learning framework. Extensive experiments conducted on both synthetic and real-world datasets unequivocally demonstrate that the proposed framework outperforms existing state-of-the-art (SoTA) joint learning frameworks and task-specific approaches in terms of both accuracy and efficiency. Additionally, the proposed framework can be seamlessly integrated with either single-modal or feature-fusion scene parsing networks, underscoring its strong versatility and broad applicability. 
In a nutshell, the main contributions of this article include: 
\begin{itemize}
    \item A general-purpose two-stream interactive joint learning framework that simultaneously addresses scene parsing and geometric vision tasks.

    \item A cross-task adapter that effectively infuses rich decoded geometric features into the scene parsing stream for improved semantic or instance segmentation.

    \item A semi-supervised training strategy tailored for geometric vision tasks based on uncertainty-aware pseudo correspondence generation.
    
    \item Extensive quantitative and qualitative experiments on multiple public datasets to validate the effectiveness and superiority of the proposed framework.
\end{itemize}

The remainder of this article is organized as follows: 
Sect. \ref{sec.related_works} reviews related previous studies.
Sect. \ref{sec.methodology} introduces the proposed framework.
Sect. \ref{sec.experiments} provides comprehensive comparisons with SoTA approaches and presents the ablation studies.
Finally, Sect.\ref{sec.conclusion} concludes this article and discusses directions for future work.

\section{Literature Review}
\label{sec.related_works}
    
\subsection{Scene Parsing Tasks}
SoTA semantic and instance segmentation networks can generally be categorized as single-modal and feature-fusion ones~\cite{dformerv2}. Early studies~\cite{bisenetv2, knet, segmenter, segformer, mask_rcnn, SparseInst} primarily adopt a single encoder to extract multi-scale contextual features from RGB images, which are subsequently decoded to produce segmentation results. Among these networks, MaskFormer~\cite{maskformer} and Mask2Former~\cite{mask2former} have notably advanced this field by unifying scene parsing tasks within a query-based prediction framework. These approaches employ a multi-scale Transformer decoder to progressively refine a set of learnable queries, enabling the generation of class-specific masks and achieving superior performance compared to conventional per-pixel classification approaches.
Nevertheless, relying on a single modality or source limits these methods from exploiting the complementary information available in other modalities or sources, often leading to degraded performance under challenging environmental conditions~\cite{sneroadseg}. To address this limitation, feature-fusion networks, such as CMX~\cite{cmxnet} and the DFormer series~\cite{dformer, dformerv2}, have gained increasing attention in recent years. These networks typically adopt duplex encoders to extract heterogeneous features from multiple modalities or sources of data and fuse these features to deliver more robust and comprehensive scene understanding.
However, the extra encoder required to process the additional data modality or source introduces substantial computational overhead. In this article, we address this issue by incorporating geometric vision tasks into a joint learning framework, which replaces the heavy encoder with a lightweight geometric vision stream that directly provides informative geometric features. This design significantly reduces the computational cost compared to previous feature-fusion networks while simultaneously improving scene parsing performance.

\subsection{Geometric Vision Tasks}
Stereo matching and optical flow estimation, two representative geometric vision tasks, aim to produce dense correspondences across images captured from different viewpoints~\cite{correspondence_tip}. Existing approaches rely either on cost volume construction or iterative refinement. The former, exemplified by PSMNet~\cite{psmnet} and DCFlow~\cite{dcflow}, typically construct a multi-dimensional cost volume that is subsequently processed using stacks of 3D convolutional layers to aggregate local and global contextual information. While highly effective, these methods often incur substantial computational and memory costs due to the extensive use of 3D convolutions~\cite{igevstereo}.
To address these limitations, iterative refinement-based methods, such as RAFT~\cite{raft} and RAFT-Stereo~\cite{raftstereo}, have emerged in recent years. These approaches generally employ gated recurrent units (GRUs) to iteratively refine the estimated correspondences through the recurrent updates of the hidden states. Recent advances have concentrated on architectural innovations within this framework. For example, FlowFormer~\cite{flowformer} expands the optical flow search range by globally aggregating correlation features via a Transformer architecture, whereas IGEV-Stereo~\cite{igevstereo} constructs a so-called combined geometry encoding volume that encodes both global and local context information for improved stereo matching. In this article, we build upon the all-pair correlation features and multi-level GRUs introduced in RAFT-Stereo~\cite{raftstereo} to develop a unified architecture capable of addressing both geometric vision tasks.

Nevertheless, the above-mentioned approaches typically require large amounts of training data with subpixel-level correspondence annotations, which are costly to obtain in real-world scenarios due to the need for highly precise LiDAR-camera calibration and data fusion~\cite{LiDAR-camera}. To mitigate this drawback, several studies~\cite{es3net,selflow,flow2stereo} formulate these two geometric vision tasks as unsupervised learning problems relying on photometric consistency. However, such photometric constraints often break down in occluded and texture-less regions, leading to substantial correspondence mismatches.
More recently, DualNet~\cite{dualnet_wang_aaai25} and Semi-Stereo~\cite{semi_stereo} introduce semi-supervised training strategies for stereo matching, where a teacher model generates pseudo labels to supervise the training of a student model. However, these approaches primarily rely on explicit cost volume filtering to obtain reliable pseudo labels, making them inapplicable to iterative refinement-based approaches. 
On the other hand, CST-Stereo~\cite{cst_stereo} introduces a consistency-aware semi-training framework tailored for iterative refinement-based stereo matching networks. Nonetheless, it requires estimating disparity maps at different resolutions, which significantly increases inference time and computational overhead. 
In this article, we introduce a semi-supervised training strategy applicable to both optical flow estimation and stereo matching based on aleatoric uncertainty estimation and adaptive pseudo label selection. The proposed strategy effectively uses large-scale, unlabeled multi-view data to improve the performance of the overall joint learning framework.

\label{stereo_related}
\subsection{Multi-Task Joint Learning}
\label{joint_related}
Existing joint learning frameworks primarily focus on leveraging one task to provide auxiliary supervision for another, with a particular emphasis on the joint learning of semantic segmentation and stereo matching.
As early efforts in this direction, DispSegNet~\cite{DispSegNet} utilizes semantic embeddings learned from the scene parsing branch to refine initial disparity estimations, while RTS$^2$Net~\cite{rts2net} adopts a multi-stage framework that exploits coarse-to-fine semantic cues to improve the performance of real-time stereo matching. Our previous work, S$^3$M-Net~\cite{s3mnet} introduces an end-to-end cascaded joint learning framework that simultaneously trains an iterative stereo matching network alongside a feature-fusion semantic segmentation network. Our subsequent study, TiCoSS~\cite{ticoss} further improves the semantic segmentation performance by tightening the coupling between these two tasks. However, such joint learning frameworks remain unidirectional, with a primary emphasis on leveraging the stereo matching branch to evolve the semantic segmentation branch. Discussions on bidirectional feature-level interactions, particularly between general scene parsing and geometric vision tasks, remain limited. Therefore, this study proposes a novel two-stream interactive joint learning framework to mitigate this gap.

Regarding the training strategy, most existing joint learning approaches rely on fully supervised learning and require both high-quality disparity and segmentation annotations, making the acquisition of well-labeled real-world datasets highly desirable yet challenging. To address this issue, DSNet~\cite{dsnet} adopts an alternant joint learning strategy, where the parameters of the semantic segmentation and stereo matching networks are alternately frozen during training on two independent datasets. Nevertheless, this decoupled training strategy limits the exploitation of shared contextual and geometric information, as the features are not jointly optimized in an end-to-end manner. While SegStereo~\cite{segstereo} and SG-RoadSeg~\cite{sgroadseg} introduce an unsupervised stereo matching network into their joint learning framework, the limited accuracy of these unsupervised approaches often leads to unreliable geometric information, which can further degrade the performance of scene parsing tasks within a joint learning framework. The proposed TwInS framework incorporates a semi-supervised training strategy for geometric vision tasks within the joint learning framework, effectively addressing the limitations posed by unreliable supervision on correspondence matching.

\begin{figure*}[!t]
	\centering
	\includegraphics[width=0.99\textwidth]{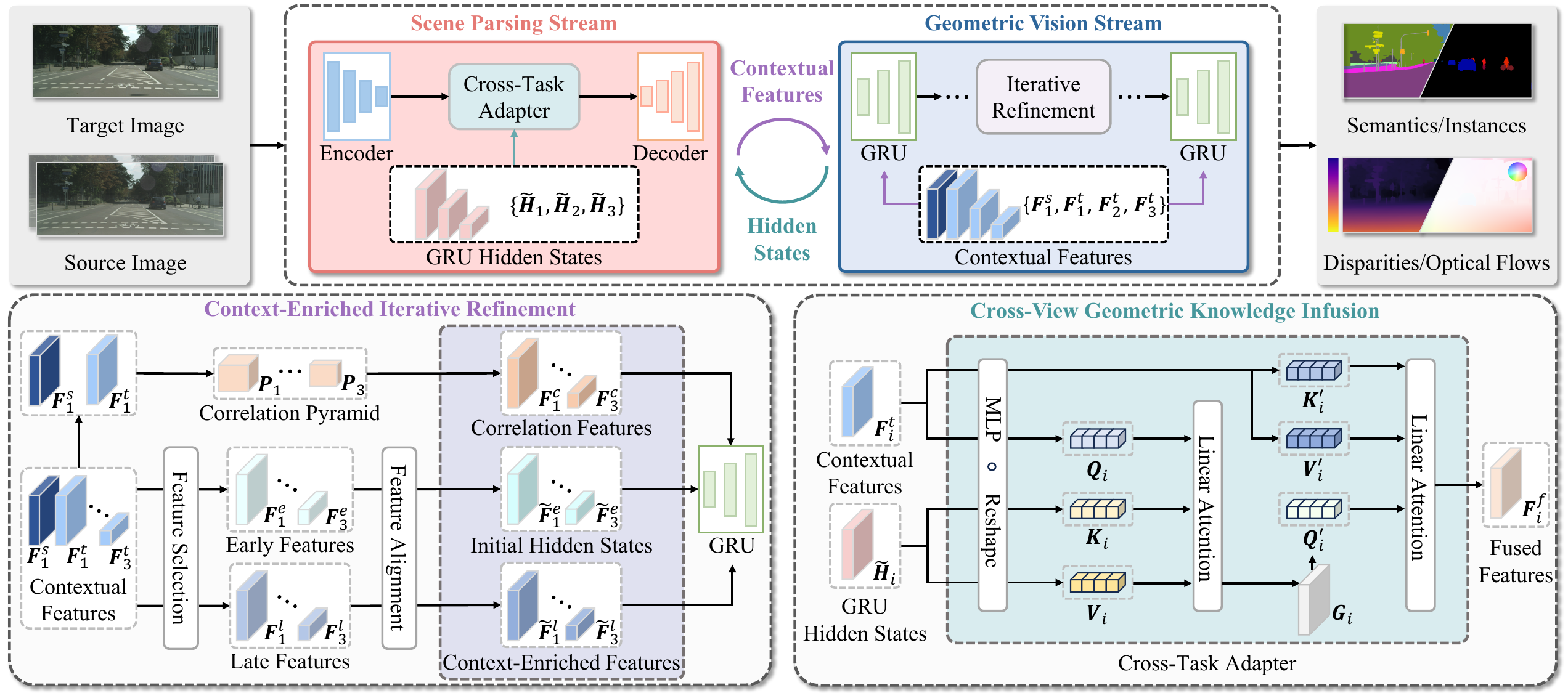}
	\caption{
    An illustration of the proposed TwInS framework A pair of stereo images or consecutive video frames are fed into the framework. Contextual features extracted by the scene parsing encoder are used to guide context-enriched iterative refinement in the geometric vision stream. The iteratively refined GRU hidden states are subsequently fed back into the scene parsing stream via a cross-task adapter which provides complementary geometric cues to enhance the scene parsing performance.
    }
    \vspace{-1.2em}
	\label{fig.method}
\end{figure*}
\section{Methodology}
\label{sec.methodology}
\subsection{Overview}
\label{workflow}
As illustrated in Fig. \ref{fig.method}, the proposed TwInS framework consists of two bidirectionally interactive streams that jointly address scene parsing and geometric vision tasks: the scene parsing stream performs semantic or instance segmentation via a unified query-based mask prediction framework, while the geometric vision stream adopts an iterative refinement strategy to estimate correspondences in a coarse-to-fine manner.
Specifically, a convolutional neural network (CNN)-based encoder is first employed to extract multi-scale contextual features from either stereo image pairs or consecutive video frames for scene parsing. These informative contextual features are then shared with the geometric vision stream 
to provide features required for updating GRUs, thereby guiding the iterative refinement process. In turn, the geometric features generated by GRUs, which capture rich cross-view geometric cues across iterations, are fed back into the scene parsing stream through a cross-task adapter to provide crucial 3D geometric information, enabling more comprehensive scene understanding. 

As for the training paradigm, a thorough search of the literature indicates that TwInS is the first joint learning framework capable of learning geometric vision tasks in a semi-supervised manner, enabling the effective exploitation of large-scale multi-view data. Specifically, a teacher model is first pre-trained on large-scale synthetic data in a fully supervised manner to simultaneously learn correspondence matching and aleatoric uncertainty estimation. The well-trained teacher model is then applied to real-world data, where an adaptive quantile-based threshold is derived from the resulting uncertainty distribution to select sparse yet reliable pseudo labels for the supervision of a student model with an identical architecture. During training, the teacher model is updated through the exponential moving average~(EMA)~\cite{dinov2} of the student model’s parameters, which progressively improve the pseudo label quality.

\subsection{Context-Enriched Iterative Refinement}
\label{sec.method1}
Existing networks for geometric vision tasks often struggle to produce accurate correspondences in texture-less or reflective regions due to the absence of discriminative photometric cues~\cite{segstereo}. In these challenging regions, features learned from scene parsing tasks can provide valuable contextual cues to alleviate such ambiguities, since correspondences typically exhibit smooth variations within the same object. Therefore, prior studies~\cite{DispSegNet, segstereo} introduced semantic segmentation as an auxiliary task and incorporated high-level contextual features into the cost computation and aggregation processes to improve correspondence estimation. Nevertheless, these previous methods are not directly applicable to current mainstream networks based on iterative refinement~\cite{raft,raftstereo}, which typically employ GRUs to progressively update correspondences. More recent studies~\cite{s3mnet,ticoss} have jointly trained iterative stereo matching networks together with semantic segmentation networks, but their employed cascaded designs fail to fully exploit the informative contextual features learned from semantic segmentation to enhance stereo matching. As a result, the interaction between the two tasks remains limited, often leading to imbalanced task optimization. Therefore, the proposed TwInS framework explicitly infuses multi-scale contextual features from the scene parsing stream into the iterative refinement process, thereby providing informative contextual cues to mitigate correspondence matching ambiguities in ill-posed regions.

As illustrated in Fig.~\ref{fig.method}, given a pair of target and source RGB images\footnote{In this article, the subscripts ``$t$'' and ``$s$'' denote ``target'' and ``source'', respectively.} $\boldsymbol{I}^{t,s} \in \mathbb{R}^{H\times W\times 3}$, where $H$ and $W$ represent the image height and width, a pre-trained CNN-based encoder ConvNeXt~\cite{convnext}, capable of capturing both local context and global semantics, is employed to extract multi-scale contextual features $\mathcal{F}_c^t=\left \{\boldsymbol{F}_1^{t},...,\boldsymbol{F}_4^{t} \right \}$ and $\mathcal{F}_c^s=\left \{ \boldsymbol{F}_1^{s},...,\boldsymbol{F}_4^{s} \right \}$ for scene parsing tasks. $\boldsymbol{F}_i^{t,s}\in \mathbb{R}^{{\frac{H}{S_i}}\times \frac{W}{S_i} \times C_i }$ denotes the features at the $i$-th encoding stage, where $C_i$ and $S_i=2^{i+1} \left(i \in \left[1,4\right] \cap \mathbb{Z}\right)$ denote the channel and stride numbers, respectively.
A context-aware correlation volume $\boldsymbol{V}$ is then constructed using $\boldsymbol{F}_1^{t}$ and $\boldsymbol{F}_1^{s}$, the features extracted at the initial encoding stage. The correlation $V_{\boldsymbol{p}, \boldsymbol{q}}$ between $\boldsymbol{p}$ and $\boldsymbol{q}$, two locations in the target and source feature maps, respectively, is computed as follows:
\begin{equation}
\label{eq:correlation}
V_{\boldsymbol{p}, \boldsymbol{q}}
= \sum_{c=0}^{C_1-1} 
\boldsymbol{F}_1^t(\boldsymbol{p}, c)\cdot
\boldsymbol{F}_1^s(\boldsymbol{q}, c).
\end{equation}
Following the study~\cite{raftstereo}, repeated average pooling is further performed on $\boldsymbol{V}$ to obtain a multi-level correlation pyramid $\mathcal{P}=\{\boldsymbol{P}_1, \boldsymbol{P}_2, \boldsymbol{P}_3\}$, which enables the network to effectively handle both large and small displacements.
Additionally, unlike previous methods~\cite{raft,raftstereo} that utilize multiple independent shallow encoders to extract context-enriched features and initialize the hidden states of GRUs, this study instead obtains these features directly from the scene parsing stream. By infusing these features into the GRUs at each iteration, the geometric vision stream is provided with rich contextual cues that guide the iterative refinement process, thereby improving the correspondence matching robustness in ambiguous or ill-posed regions.
Specifically, each encoding stage consists of multiple convolutional blocks. The TwInS framework retains the outputs from both the first and last convolutional blocks of the first three encoding stages, denoted as early features $\boldsymbol{F}_i^{e}$ (from shallow layers) and late features $\boldsymbol{F}_i^{l}$ (from deep layers), respectively. The early features, rich in local textural details, are used to initialize the hidden states of the GRUs, while the late features, containing global semantic information, serve as the context-enriched features to provide high-level guidance throughout the iterative refinement process. To facilitate effective feature fusion between the two streams, the channel dimensions of the features are aligned through the following expression:
\begin{equation}
\label{eq:feature_alignment}
\fixedtilde{\boldsymbol{F}}{i}{e,l}= \mathrm{ReLU}\left(\mathrm{GroupNorm}\left(\mathrm{Conv}\left(\boldsymbol{F}_i^{e,l}\right)\right)\right),
\end{equation}
where $\mathrm{ReLU}$ is the rectified linear unit activation function, $\mathrm{GroupNorm}$ denotes the group normalization operation, and $\mathrm{Conv}$ represents a $1\times1$ convolutional layer. By combining the aligned early and late features, the GRUs are enriched with both fine-grained local details and high-level global semantics, ultimately achieving more accurate correspondence estimation. 

\subsection{Cross-View Geometric Knowledge Infusion}
\label{sec.method2}

The previous cascaded joint learning framework~\cite{s3mnet} feeds the disparities estimated using an iterative stereo matching network into a feature-fusion semantic segmentation network, thereby tightly coupling the two networks at the output level. 
Despite achieving impressive results, it often produces unreliable geometric features in regions with inaccurate disparity estimations, which can further mislead semantic segmentation within a unified joint learning framework~\cite{roadformer}. To alleviate this issue, the study~\cite{ticoss} introduces a gated feature fusion strategy to filter out unreliable features. Nonetheless, this approach still fails to provide effective geometric feature guidance for the semantic segmentation network in these ambiguous regions.   
Additionally, existing methods often incur significant computational overhead due to the use of duplex encoders, an inherent limitation of feature-fusion networks, which restricts their deployment on resource-constrained hardware.

In this article, the geometric features generated by GRUs, capable of capturing cross-view geometric cues through iterative refinement, are utilized to provide more reliable and informative geometric information to the scene parsing stream.
Specifically, within the geometric vision stream, iterative refinement is performed by recurrently updating the multi-level hidden states, which can be formulated as follows:
\begin{equation}
\label{eq:update_gru}
\boldsymbol{H}_j^{k+1}=\mathrm{GRU}(\boldsymbol{H}_j^{k}, \fixedtilde{\boldsymbol{F}}{j}{l}, \boldsymbol{F}_{j}^{c}),
\end{equation}
where $\boldsymbol{H}_j^{k}$ denotes the GRU hidden states at update level $j$ during the $k$-th iteration, $\fixedtilde{\boldsymbol{F}}{j}{l}$ represents the contextual features aligned through \eqref{eq:feature_alignment}, and $\boldsymbol{F}_{j}^{c}$ is the correlation features derived from $\boldsymbol{P}_{j}$. The hidden states across all iterations are subsequently processed by multiple convolutional layers to directly regress correspondences, resulting in a sequence of progressively refined outputs. These hidden states primarily capture coarse global geometric features in the early iterations, where correspondence updates involve large displacements. As the optimization converges and the magnitude of updates decreases, the hidden states progressively shift the focus toward fine-grained local geometric details. As a result, the final hidden states retain rich geometric information accumulated throughout the entire iterative correspondence matching process. Compared to features obtained by directly encoding the disparities or optical flows, these iteratively refined hidden states provide more comprehensive and informative geometric cues for the scene parsing stream.

Nevertheless, a significant domain gap exists between the hidden states learned for geometric vision tasks and the contextual features learned for scene parsing tasks, owing to their inherently different optimization objectives. Previous feature fusion methods~\cite{s3mnet,ticoss,roadformer} tailored for heterogeneous features extracted from duplex encoders are not directly applicable to fusing such task-specific intermediate features. To address this limitation, a CTA is further designed within the TwInS framework to enable effective cross-task knowledge infusion from the geometric vision stream into the scene parsing stream. Specifically, channel and resolution alignment is first performed between the hidden states and the contextual features, resulting in a set of aligned hidden states $\mathcal{H}=\{\tilde{\boldsymbol{H}}_1,...,\tilde{\boldsymbol{H}}_3\}$. Inspired by the recent work~\cite{Rad2021ICML} that performs cross-modal feature alignment in a shared embedding space, the proposed CTA first projects the hidden states into the contextual feature space to obtain complementary geometric cues using the following expression:
\begin{equation}
\label{eq:linear_attention}
\begin{aligned}
\boldsymbol{G}_i =\mathrm{MLP}(\mathrm{RMSNorm}(\frac{\phi(\boldsymbol{Q}_i)(\phi(\boldsymbol{K}_i)^{^\top} {\boldsymbol{V}_i})}
{\phi(\boldsymbol{Q}_i) (\mathrm{Reshape(}\phi({\boldsymbol{K}}_{i})^\top))})),
\end{aligned}
\end{equation}
where $\boldsymbol{G}_i \in \mathbb{R}^{{\frac{H}{S_i}}\times \frac{W}{S_i} \times C_i}$ denotes the projected geometric features at the $i$-th stage, $\mathrm{MLP}$ represents a multi-layer perceptron, $\mathrm{RMSNorm}$ refers to the root-mean-square layer normalization, $\phi(\cdot)$ denotes an ELU activation function with a constant shift, enabling the decomposition of the softmax-based attention into a linear form to reduce the computation complexity~\cite{linear_attention}, the query embeddings $\boldsymbol{Q}_i\in \mathbb{R}^{{\frac{H}{S_i}}\times \frac{W}{S_i}\times C_i}$ are derived from $\boldsymbol{F}_i^t$, the key and value embeddings, $\boldsymbol{K}_i$ and $\boldsymbol{V}_i$, are derived from $\tilde{\boldsymbol{H}}_i$, and $\mathrm{Reshape}$ represents the dimension transformation operation.
The multi-level contextual features are remapped as query embeddings to adaptively retrieve task-relevant geometric cues from the aligned hidden states via~\eqref{eq:linear_attention}, thereby preventing the introduction of interfering noise into the scene parsing stream. After obtaining the complementary geometric features, the next objective is to fuse them with the contextual features to deliver more robust and comprehensive scene parsing. To this end, the projected geometric features $\boldsymbol{G}_i$ (remapped as query embeddings $\boldsymbol{Q}_i^{\prime}$) are further utilized to selectively aggregate the contextual features $\boldsymbol{F}_i^t$ (remapped as key and value embeddings, $\boldsymbol{K}_i^{\prime}$ and $\boldsymbol{V}_i^{\prime}$, respectively) via the linear attention mechanism. This process yields multi-level fused features {$\boldsymbol{F}_i^f$} that incorporate both contextual and geometric cues. The final scene parsing results are subsequently generated by a unified query-based mask prediction decoder. By leveraging the GRU hidden states from the correspondence matching network, valuable cross-view geometric knowledge is effectively infused into the scene parsing stream with negligible computational overhead. As demonstrated by the t-distributed stochastic neighbor embedding (t-SNE)~\cite{tsne_2008_jmlr} visualization results presented in Sect.~\ref{sec.exp_ablation}, these fused features exhibit improved intra-class compactness and inter-class separability in high-dimensional feature space.

\subsection{Semi-Supervised Training Based on Uncertainty-Aware Pseudo Correspondence Generation}
\label{sec.method3}

Regarding the training strategy, existing joint learning frameworks typically adopt either supervised or unsupervised learning methods to train the stereo matching branch~\cite{ticoss,sgroadseg}. However, supervised approaches require sub-pixel level disparity annotations, which are scarce in large-scale driving scene parsing datasets. In contrast, unsupervised learning approaches rely on photometric consistency, which is inherently fragile in occluded or texture-less regions, ultimately compromising the accuracy of semantic segmentation within the joint learning framework. Inspired by recent knowledge distillation-based self-training strategies employed to train vision foundation models~\cite{dinov2,depthanything}, which leverage a pre-trained teacher model to generate pseudo labels for supervising a student model on large-scale unlabeled data, several studies~\cite{rose_wang_tcsvt25,dualnet_wang_aaai25} have adopted similar strategies to train stereo matching models. Within this knowledge distillation framework, the student model's performance heavily depends on the quality of the pseudo labels. Nevertheless, existing approaches commonly rely on explicit cost volume filtering combined with left-right consistency checks to obtain these pseudo labels, which is not only computationally expensive but also incompatible with iterative refinement architectures. 

To address this limitation, this study introduces a novel semi-supervised training strategy tailored for geometric vision tasks based on aleatoric uncertainty estimation and adaptive pseudo label selection. Specifically, a teacher model is first pre-trained in a fully supervised manner on large-scale synthetic data, where both segmentation and correspondence annotations are accessible.
According to the analysis in the study~\cite{kendall_2017_uncertainties}, when a correspondence matching model is trained by minimizing the mean absolute error (MAE) loss, the prediction residuals can be modeled as samples from a zero-mean Laplace distribution, leading to the following likelihood formulation:
\begin{equation}
\label{eq:laplace}
p(\boldsymbol{x}_{c} \mid \hat{\boldsymbol{x}}_{c},{\sigma})
= \frac{1}{2{\sigma}}
\exp\!\left(
 -\frac{\left\|\boldsymbol{x}_{c} - \hat{\boldsymbol{x}}_{c}\right\|_1}{{\sigma}}
\right),
\end{equation}
where $\boldsymbol{x}_{c}\in \mathbb{R}^2$ denotes the ground-truth correspondence\footnote{In this article, for notational simplicity, disparity is also formulated as a two-dimensional vector, with the second component fixed to zero.} at a given pixel in the target image, and $\hat{\boldsymbol{x}}_{c}\in \mathbb{R}^2$ denotes the correspondence produced by the model. The scale parameter $\sigma$ controls the spread of the Laplace distribution, where a smaller value indicates a higher concentration of probability mass around the zero-mean residual, corresponding to higher prediction confidence and lower aleatoric uncertainty.

The previous work~\cite{sednet_chen_cvpr23} estimates such uncertainties by computing pair-wise differences among multi-resolution disparity maps, which are, nevertheless, inaccessible in the TwInS framework, due to its reliance on iterative refinement rather than multi-scale decoding. In contrast, this study derives uncertainties from a set of iterative correspondences $\{\fixedhat{\boldsymbol{X}}{c}{1},...,\fixedhat{\boldsymbol{X}}{c}{K}\}$, where $K$ represents the total number of refinement iterations. Intuitively, correspondences that exhibit persistent fluctuations across iterations are indicative of higher uncertainty. An uncertainty map $\boldsymbol{U}$, with each value $\sigma$ representing the predicted uncertainty at a given pixel, can be obtained by analyzing the pixel-wise fluctuations among iteratively matched correspondences using the following expression:
\begin{equation}
{\boldsymbol{U}} =\mathrm{MLP}\left(\mathrm{Concat}\left(\left\{\,  \psi(\fixedhat{\boldsymbol{X}}{c}{i} - \fixedhat{\boldsymbol{X}}{c}{j}) \,\middle|\, 1 \leq i<j \leq K \right\}\right)\right),
\label{eq:pdv_uncertainty}
\end{equation}
where $\mathrm{Concat}$ denotes the channel-wise concatenation operation, and $\psi(\cdot)$ is an element-wise square function. Following the study~\cite{sednet_chen_cvpr23}, this work minimizes the negative log-likelihood of~\eqref{eq:laplace} to learn pixel-wise uncertainty estimation, and further aligns the uncertainty distribution with the Laplace distribution of the prediction residuals by minimizing the Kullback–Leibler divergence.

After being pre-trained on large-scale synthetic data, the teacher model is applied to unlabeled real-world data to generate pseudo labels for a student model with an identical architecture. To filter out unreliable correspondences, a quantile-based threshold $\tau$ is computed for each uncertainty map using the Laplace quantile function expressed as follows:
\begin{equation}
\tau = \mu + b \cdot \log\left(2(1 - \alpha)\right),
\label{eq:adaptive_threshold}
\end{equation}
where \textbf{$\mu$} and $b$ represent the median and the mean absolute deviations estimated from the uncertainty map, respectively, and $\alpha$ is a hyper-parameter specifying the desired confidence percentile. The impact of $\alpha$ on the overall framework performance is further discussed in Sect.~\ref{sec.exp_ablation}. The threshold is subsequently applied to the uncertainty map to identify reliable correspondences from the teacher model, wherein pixels with uncertainty values lower than $\tau$ are selected as pseudo labels to supervise the training of the student model. Compared to using a fixed threshold, the quantile-based strategy adaptively determines an image-specific threshold based on estimated uncertainties, thereby enabling more flexible and robust pseudo label selection across diverse environments.
Additionally, during training, the teacher model is provided with weakly augmented images to generate reliable pseudo labels, while the student model is trained on strongly augmented images and encouraged to produce consistent predictions under greater perturbations. The teacher model's parameters are updated using the EMA strategy~\cite{dinov2} to progressively improve the pseudo label quality. As demonstrated by the ablation studies in Sect.~\ref{sec.exp_ablation}, the proposed semi-supervised training strategy enables effective optimization of the TwInS framework on large-scale scene parsing datasets, thereby significantly improving its performance on both scene parsing and geometric vision tasks.

\section{Experiments}

\begin{figure*}[!t]
	\centering
	\includegraphics[width=0.99\textwidth]{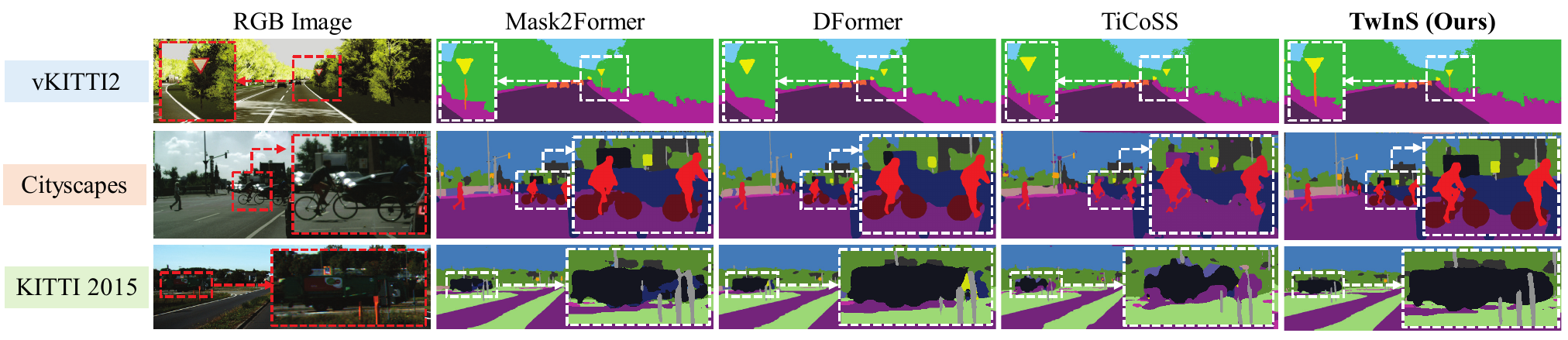}
	\caption{Qualitative results of SoTA semantic segmentation networks on the vKITTI2, Cityscapes, and KITTI 2015 datasets.}
	\label{fig.semseg}
    \vspace{-1.0em}
\end{figure*}

\begin{figure*}[!t]
	\centering
	\includegraphics[width=0.99\textwidth]{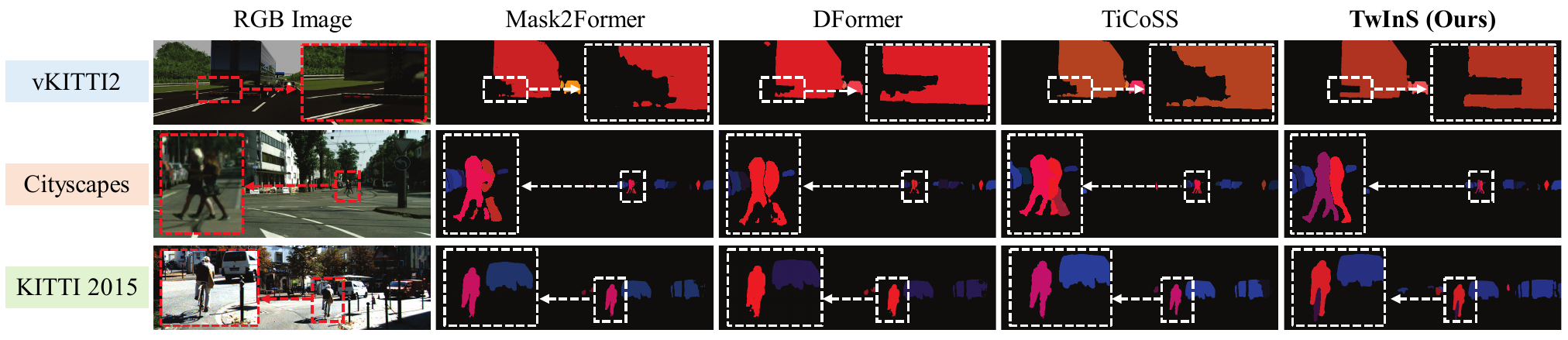}
	\caption{Qualitative results of SoTA instance segmentation networks on the vKITTI2, Cityscapes, and KITTI 2015 datasets.}
	\label{fig.insseg}
    \vspace{-0.5em}
\end{figure*}

\begin{table*}[!t]
    \settablefont
	\centering
	\caption{
    Quantitative comparisons of SoTA semantic segmentation and instance segmentation networks on the VKITT2, Cityscapes, and KITTI 2015 datasets. 
	}
    \label{tab.seg_overall}
    
    \begin{subtable}[t]{\textwidth}
    \centering
    \setlength{\tabcolsep}{1.5mm}
    \begin{tabular}
	    {L{2.2cm}|C{1.8cm}|C{2.6cm} |C{1.45cm} C{1.45cm} | C{1.45cm} C{1.45cm} | C{1.45cm} C{1.45cm}}
		\toprule
        \multirow{2}{*}{Networks} & \multirow{2}{*}{Publication} & \multirow{2}{*}{Type} 
        & \multicolumn{2}{c}{vKITTI2} 
        & \multicolumn{2}{c}{Cityscapes} 
        & \multicolumn{2}{c}{KITTI 2015}
        \\
        \cline{4-9}
        & & 
        & mIoU (\%) $\uparrow$ & mFSc (\%) $\uparrow$ 
        & mIoU (\%) $\uparrow$ & mFSc (\%) $\uparrow$ 
        & mIoU (\%) $\uparrow$ & mFSc (\%) $\uparrow$ \\
		\hline
		BiSeNet V2~\cite{bisenetv2} & IJCV'21 &\multirow{5}{*}{Single-Modal Networks} &70.41 &80.83 &51.72 &62.91 &40.78 &58.20 \\
		Segmenter~\cite{segmenter} & ICCV'21 & &83.53 &90.22 &72.88 &83.44 &61.47 &79.57 \\
		SegFormer~\cite{segformer} & NeurIPS'21 & &88.02 &93.34 &60.13 &85.60 &67.84 &81.39 \\
		KNet~\cite{knet} & NeurIPS'21 & &85.05 &91.45 &72.78 &83.08 &60.13 &73.82 \\
		Mask2Former~\cite{mask2former} & CVPR'22 & &86.83 &92.65 &78.41 &86.29 &67.48 &78.50 \\
        \hline
        CMX~\cite{cmxnet} & TITS'23 &\multirow{5}{*}{Feature-Fusion Networks} &86.01 &92.04 &73.55 &83.76 &60.68 &71.75 \\
        DFormer~\cite{dformer} & ICLR'24 & &85.99 &89.37 &77.76 &86.89 &70.13 &80.43 \\
        RoadFormer~\cite{roadformer} & TIV'24 & &87.75 &93.45 &76.82 &86.11 &65.51 &79.59 \\
        RoadFormer+~\cite{roadformer+} & TIV'24 & &88.44 &93.52 &80.73 &88.83 &70.90 &79.54 \\
        DFormerv2~\cite{dformer} & CVPR'25 & &81.99 &86.98 &75.56 &85.36 &70.45 &78.61 \\
        \hline
        DINO V2~\cite{dinov2} & ICLR'23 &\multirow{3}{*}{Vision Foundation Models} &86.36 &92.18 &82.49 &89.99 &74.65 &84.14 \\
        Depth Anything~\cite{depthanything} & CVPR'24 & &92.10 &96.27 &82.98 &90.31 &73.82 &83.21 \\
        Vit-CoMer~\cite{vit_comer} & CVPR'24 & &92.37 &95.93 &79.72 &88.25 &71.99 &81.81 \\
        \hline
        DSNet~\cite{dsnet} & ICRA'19 &\multirow{7}{*}{Joint Learning Approaches} &76.61 &85.36 &54.30 &59.96 &39.19 &46.11 \\
        SGDepth~\cite{SGDepth} & ECCV'20 & &80.19 &85.91 &55.96 &61.11 &40.20 &47.51 \\
        SG-RoadSeg~\cite{sgroadseg} & ICRA'24 & &81.06 &86.49 &56.37 &68.47 &32.89 &39.49 \\
        S$^3$M-Net~\cite{s3mnet} & TIV'24 & &84.00 &88.27 &58.29 &65.45 &41.54 &48.02 \\
        TiCoSS~\cite{ticoss} & TASE'25 & &87.94 &90.52 &63.57 &74.06 &47.66 &54.69 \\
        SemStereo~\cite{semstereo_chen_aaai2025} & AAAI'25 & &83.69 &87.01 &59.86 &68.42 &45.89 &52.04 \\
        \textbf{Ours} & / & &\textbf{88.67} &\textbf{93.69} &\textbf{82.13} &\textbf{89.73} &\textbf{72.36} &\textbf{82.69} \\
		\bottomrule
    \end{tabular}
    \vspace{3pt}
    \caption{Semantic Segmentation}
    \label{subtab:semseg}
    \end{subtable}

    \begin{subtable}[t]{\textwidth}
    \centering
    \setlength{\tabcolsep}{1.5mm}
    \begin{tabular}
	    {L{2.2cm}|C{1.8cm}|C{2.6cm} |C{1.45cm} C{1.45cm} | C{1.45cm} C{1.45cm} | C{1.45cm} C{1.45cm}}
		\toprule
        \multirow{2}{*}{Networks} & \multirow{2}{*}{Publication} & \multirow{2}{*}{Type} 
        & \multicolumn{2}{c}{vKITTI2} 
        & \multicolumn{2}{c}{Cityscapes} 
        & \multicolumn{2}{c}{KITTI 2015}\\
        \cline{4-9}
        & & 
        & mAP (\%) $\uparrow$ & AP50 (\%) $\uparrow$ 
        & mAP (\%) $\uparrow$ & AP50 (\%) $\uparrow$ 
        & mAP (\%) $\uparrow$ & AP50 (\%) $\uparrow$ \\
        \hline
        Mask R-CNN~\cite{mask_rcnn} & ICCV'17 &\multirow{4}{*}{Single-Modal Networks} &74.73 &82.90 &22.97 &42.19 &19.01 &32.86 \\
        MaskFormer~\cite{maskformer} & NeurIPS'21 & &76.01 &87.93 &25.40 &46.00 &20.04 &33.65 \\
        SparseInst~\cite{SparseInst} & CVPR'22 & &75.40 &88.49 &25.01 &45.51 &18.10 &32.19 \\
        Mask2Former~\cite{mask2former} & CVPR'22 & &80.79 &93.01 &30.06 &49.52 &22.79 &36.94 \\
        \hline
        CalibNet~\cite{calibnet} & TIP'24 & Feature-Fusion Network &81.91 &93.97 &29.19 &47.08 &21.63 &35.49 \\
        \hline
        \textbf{Ours} & / & Joint Learning Approach &\textbf{83.20} &\textbf{95.00} &\textbf{32.10} &\textbf{51.31} &\textbf{25.90} &\textbf{38.91} \\
		\bottomrule
    \end{tabular}
    \vspace{3pt}
    \caption{Instance Segmentation}
    \label{subtab:insseg}
    \end{subtable}
 \vspace{-2.5em}
\end{table*}

\begin{figure*}[!t]
	\centering
	\includegraphics[width=0.99\textwidth]{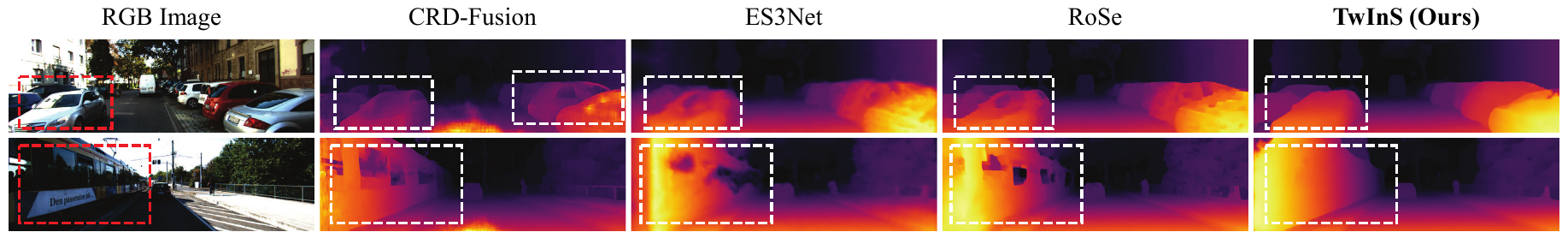}
	\caption{Qualitative results of SoTA un/semi-supervised stereo matching networks on the KITTI 2015 dataset.}
	\label{fig.stereo}
    \vspace{-1.0em}
\end{figure*}

\begin{figure*}[!t]
	\centering
	\includegraphics[width=0.99\textwidth]{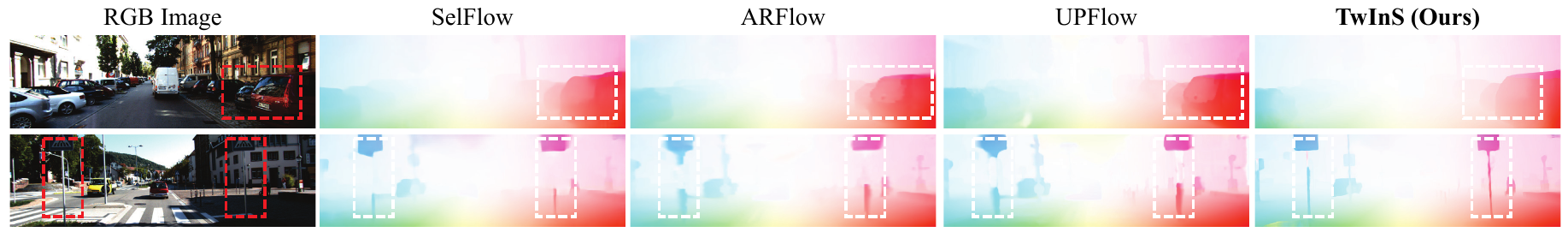}
	\caption{Qualitative results of SoTA un/semi-supervised optical flow estimation networks on the KITTI 2015 dataset.}
	\label{fig.flow}
    \vspace{-0.5em}
\end{figure*}

\label{sec.experiments}
This section presents extensive experiments conducted to evaluate the effectiveness and performance of the proposed TwInS framework. The following subsections detail the datasets, experimental setup, evaluation metrics, ablation studies, and comprehensive comparisons with previous SoTA approaches in both scene parsing and geometric vision tasks.

\subsection{Datasets}
\label{sec.datasets}
The proposed TwInS framework is pre-trained on synthetic data with both segmentation and correspondence annotations, and fine-tuned on real-world data with only segmentation annotations. The used datasets are detailed as follows:

\begin{itemize}
\item 
The vKITTI2~\cite{vkitti2} dataset provides photorealistic virtual replicas of five KITTI sequences~\cite{kitti2015}, annotated with 15 semantic classes and three instance classes. Dense ground-truth disparity and optical flow maps are rendered using a virtual engine. 
Following previous works~\cite{s3mnet,ticoss}, this study utilizes 700 collections of images to pre-train the TwInS framework in a fully supervised manner. Each collection consists of a stereo image pair along with an additional temporally successive left image. 500 collections are used for model training and the remaining 200 collections are used for model validation.
\item 
The Cityscapes~\cite{cityscapes} dataset contains extensive real-world stereo images captured in urban driving scenes, with annotations for 19 semantic classes and eight instance classes, yet no correspondence annotations. In this study, 2,975 image collections are used for model training, and 500 image collections are used for model validation. The configuration of the image collections adheres to that employed for the vKITTI2 dataset.
\item The KITTI 2015~\cite{kitti2015} dataset is widely used for benchmarking both scene parsing and geometric vision tasks. It provides segmentation annotations that are consistent with those of the Cityscapes dataset. Additionally, sparse disparity ground truth is obtained using a Velodyne HDL-64E LiDAR, while sparse optical flow ground truth is generated by fitting geometrically accurate CAD models to moving 3D point clouds derived from laser scans. These correspondence annotations are utilized to validate the effectiveness of the proposed semi-supervised training strategy for geometric vision tasks.
In this study, 140 image collections are used for model training, and 60 image collections are used for model validation. The configuration of the image collections adheres to that employed for the vKITTI2 dataset.
\end{itemize}

\begin{table}[t!]
    \settablefont
    \centering
    \caption{
    Quantitative comparisons of un/semi-supervised stereo matching and optical flow estimation networks on the KITTI 2015 dataset. 
    }
    \label{tab.geo_overall}

    \begin{subtable}[t]{\columnwidth}
        \centering
        \setlength{\tabcolsep}{1.5mm}
        \begin{tabular}{L{1.7cm}|C{1.1cm}|C{1.65cm}|C{1.45cm} C{1.3cm}}
            \toprule
            Networks & Publication & Type & EPE (pixels) $\downarrow$ & D1 (\%) $\downarrow$ \\
            \hline
            SegStereo~\cite{segstereo}     & ECCV'18 &\multirow{3}{*}{Unsupervised} & 1.59 & 7.44 \\
            CRD-Fusion~\cite{CRD-Fusion}  & CRV'22   & & 1.37 & 5.49 \\
            ES$^3$Net~\cite{es3net}       & CVPR'23  & & 1.84 & 8.79 \\
            \hline
            RoSe~\cite{rose_wang_tcsvt25} & TCSVT'25 &\multirow{2}{*}{Semi-Supervised} & 1.04 & 3.62 \\
            \textbf{Ours}                 & /       & & \textbf{0.96} & \textbf{3.34} \\
            \bottomrule
        \end{tabular}
        \vspace{3pt}
        \caption{Stereo Matching}
        \label{subtab:stereo}
    \end{subtable}

    \begin{subtable}[t]{\columnwidth}
        \centering
        \setlength{\tabcolsep}{1.5mm}
        \begin{tabular}{L{1.7cm}|C{1.1cm}|C{1.65cm}|C{1.45cm} C{1.3cm}}
            \toprule
            Networks & Publication & Type & EPE (pixels) $\downarrow$ & D1 (\%) $\downarrow$ \\
            \hline
            SelFlow~\cite{selflow} & CVPR'19 &\multirow{4}{*}{Unsupervised} & 5.12 & 15.88 \\
            ARFlow~\cite{ARflow}  & CVPR'20 & & 4.01 & 13.64 \\
            UPFlow~\cite{upflow}  & CVPR'21 & & 3.90 & 12.89 \\
            SemARFlow~\cite{semarflow}  & ICCV'23 & & 3.88 & 11.40 \\
            \hline
            \textbf{Ours}         & /    &Semi-Supervised   & \textbf{3.72} & \textbf{10.86} \\
            \bottomrule
        \end{tabular}
        \vspace{3pt}
        \caption{Optical Flow Estimation}
        \label{subtab:opflow}
    \end{subtable}
\vspace{-3.0em}
\end{table}

\subsection{Experimental Setup}
\label{sec.experimental_setup}
All experiments are conducted on two NVIDIA RTX 4090 GPUs with a total batch size of 2. Following previous studies~\cite{s3mnet, roadformer}, images from the vKITTI2 and KITTI 2015 datasets are cropped to $512 \times 256$ pixels, while the images from the Cityscapes dataset are cropped to $1024 \times 512$ pixels before being fed into the network during training.
The AdamW optimizer is employed for model training, with epsilon and weight decay set to $10^{-8}$ and $10^{-5}$, respectively. The initial learning rate is fixed at $1\times 10^{-4}$. The model is trained for 50,000 iterations on the vKITTI2 dataset, 75,000 iterations on the Cityscapes dataset, and 5,000 iterations on the KITTI 2015 dataset. Standard data augmentation techniques, including random cropping, color jittering, and asymmetric occlusion, are applied during training to improve model robustness.
\subsection{Evaluation Metrics}
\label{sec.evaluation_metrics}
This study quantifies the semantic segmentation performance using mean intersection over union (mIoU) and mean F1-score (mFSc), and the instance segmentation performance using mean average precision (mAP) and average precision at an IoU threshold of 50\% (AP50).
Moreover, this study quantifies the performance of stereo matching and optical flow estimation using average end-point error (EPE) and 
the percentage of pixels with an error exceeding both three pixels and 5\% of the ground-truth correspondences (D1).

\subsection{Comparison with SoTA Approaches}
\label{sec.SoTA_comparisions}
\subsubsection{Scene Parsing Tasks}
\label{sec.semntic_performance}
The quantitative experimental results for joint learning of stereo matching with semantic or instance segmentation across all three datasets are reported in Table~\ref{tab.seg_overall}, and the qualitative experimental results are presented in Figs.~\ref{fig.semseg} and \ref{fig.insseg}.

These results suggest that the proposed TwInS framework consistently outperforms existing single-modal, feature-fusion, and joint learning approaches across all three datasets on both scene parsing tasks. Notably, it achieves substantial performance gains over the SoTA baseline model Mask2Former, with mIoU improvements ranging from $2.12\%$ to $7.23\%$ in semantic segmentation and mAP increases from $2.89\%$ to $13.64\%$ in instance segmentation. These advancements are primarily attributed to the integration of complementary cross-view geometric cues from the geometric vision stream, which effectively resolve semantic or instance ambiguities in challenging regions such as object boundaries and occlusions, as exemplified in the third row of Fig.~\ref{fig.semseg} as well as the first row of Fig.~\ref{fig.insseg}. Remarkably, TwInS delivers superior performance even when compared with recent vision foundation models~\cite{dinov2,depthanything,vit_comer} pre-trained on extensive well-annotated data, and outperforms ViT-CoMer on both real-world datasets, underscoring its superiority in real-world driving scenarios.

Moreover, compared to SoTA feature-fusion networks, the TwInS framework achieves up to a $2.05\%$ improvement in mIoU over RoadFormer+ and up to a $19.74\%$ gain in mAP over CalibNet across the three datasets. Its superiority is further illustrated in the second row of Fig.~\ref{fig.insseg}, which presents a challenging scenario involving occluded instances of the same semantic category. In this case, only TwInS effectively distinguishes individual objects, demonstrating strong robustness in complex driving scenes. These performance gains primarily stems from the use of iteratively refined GRU hidden states in the geometric vision stream, which provide richer and more informative geometric features than those extracted directly from the disparities or optical flows.

Finally, the quantitative experimental results further demonstrate that the TwInS framework significantly surpasses previous joint learning approaches, including those leveraging task-independent decoder designs~\cite{dsnet,SGDepth} and cascaded designs, with improvements in mIoU by $29.19\%$ and $51.82\%$ over TiCoSS on the two real-world datasets.
This substantial performance gain is largely attributed to the developed semi-supervised training strategy, which effectively exploits large-scale multi-view data to improve scene parsing performance. 

Additional comparisons against SoTA joint learning methods on the KITTI Semantics benchmark, along with further qualitative results across the three datasets, are provided in the supplement.

\subsubsection{Geometric Vision Tasks}
\label{sec.exp_disp}
The quantitative experimental results for joint learning of semantic segmentation with stereo matching or optical flow estimation on the KITTI 2015 dataset are provided in Table~\ref{tab.geo_overall}, and the qualitative results are shown in Figs.~\ref{fig.stereo} and \ref{fig.flow}. 
Specifically, the TwInS framework is compared with both unsupervised and semi-supervised approaches with official implementations that are publicly available. As reported, the TwInS framework achieves considerable performance gains, reducing the EPE by up to $47.82\%$ for stereo matching and $37.63\%$ for optical flow estimation. Notably, it is observed that semi-supervised approaches, including RoSe and TwInS, consistently outperform unsupervised approaches. This performance gap primarily stems from the inherent limitation of the photometric consistency loss, which often fails to provide effective supervisory signals, particularly in texture-less and reflective regions. In contrast, the semi-supervised training strategy adopted in the TwInS framework leverages explicit pixel-wise constraints to provide more direct and reliable supervision, enabling the generation of more accurate correspondences in these challenging regions.

\begin{table}
        \settablefont
	\centering
    \caption{
    Ablation study on the context-enriched iterative refinement in the geometric vision stream conducted on the KITTI 2015 dataset.  
    }
\label{tab.context_enriched}
	\setlength{\tabcolsep}{1.5mm}
	\begin{tabular}
	    {C{1.3cm}| C{1.3cm}| C{1.3cm}| C{1.55cm} C{1.55cm}}
		\toprule
            Correlation  & Context & Hidden  & \multirow{2}{*}{EPE (pixels) $\downarrow$}  & \multirow{2}{*}{D1 (\%) $\downarrow$} \\
            Pyramid & Features & States & & \\
		\hline    \hline
            \multicolumn{3}{c|}{Baseline} &1.13      &4.08         \\
            \hline
            \checkmark & &   &1.10     &3.96  \\
            \checkmark &\checkmark &  &1.06     &3.90    \\
            \checkmark &\checkmark &\checkmark   &\textbf{1.00}  &\textbf{3.84} \\
            
            \bottomrule
		\end{tabular}
         \vspace{-1.5em}
\end{table}

\begin{figure}[!t]
    \centering
    
    \begin{minipage}[b]{0.5\textwidth}
        \settablefont
        \centering
        \captionof{table}{
            Ablation study on the feature fusion strategy conducted on the Cityscapes and KITTI 2015 datasets. The proposed cross-task adapter is compared with two SoTA feature fusion strategies that incorporate cross-view geometric cues into the scene parsing stream.
        }
        \label{tab.cross_task_adapter}
        \setlength{\tabcolsep}{1.5mm}
        \begin{tabular}{L{1.5cm}|C{1.2cm}C{1.2cm}|C{1.2cm}C{1.2cm}}
        \toprule
        {\multirow{2}{*}{Methods}} &  
        \multicolumn{2}{c}{Cityscapes} & 
        \multicolumn{2}{c}{KITTI 2015} \\
        \cline{2-5}
        & mIoU (\%) $\uparrow$ & mAcc (\%) $\uparrow$ & mIoU (\%) $\uparrow$ & mAcc (\%) $\uparrow$ \\
        \hline
        \hline
        Baseline  &78.52  &86.80 &68.19 &78.54  \\
        FFM~\cite{s3mnet} &78.81  &87.36 &67.81 &79.02   \\
        HFSB~\cite{roadformer} &76.42 &86.02 &66.06 &77.23 \\
        CTA \textbf{(Ours)} &\textbf{80.05}  &\textbf{87.98} &\textbf{72.03} &\textbf{80.92}  \\ 
        \bottomrule
        \end{tabular}
        \vspace{-1.5em}
    \end{minipage}
    
    \vspace{1cm}
    
    \includegraphics[width=0.48\textwidth]{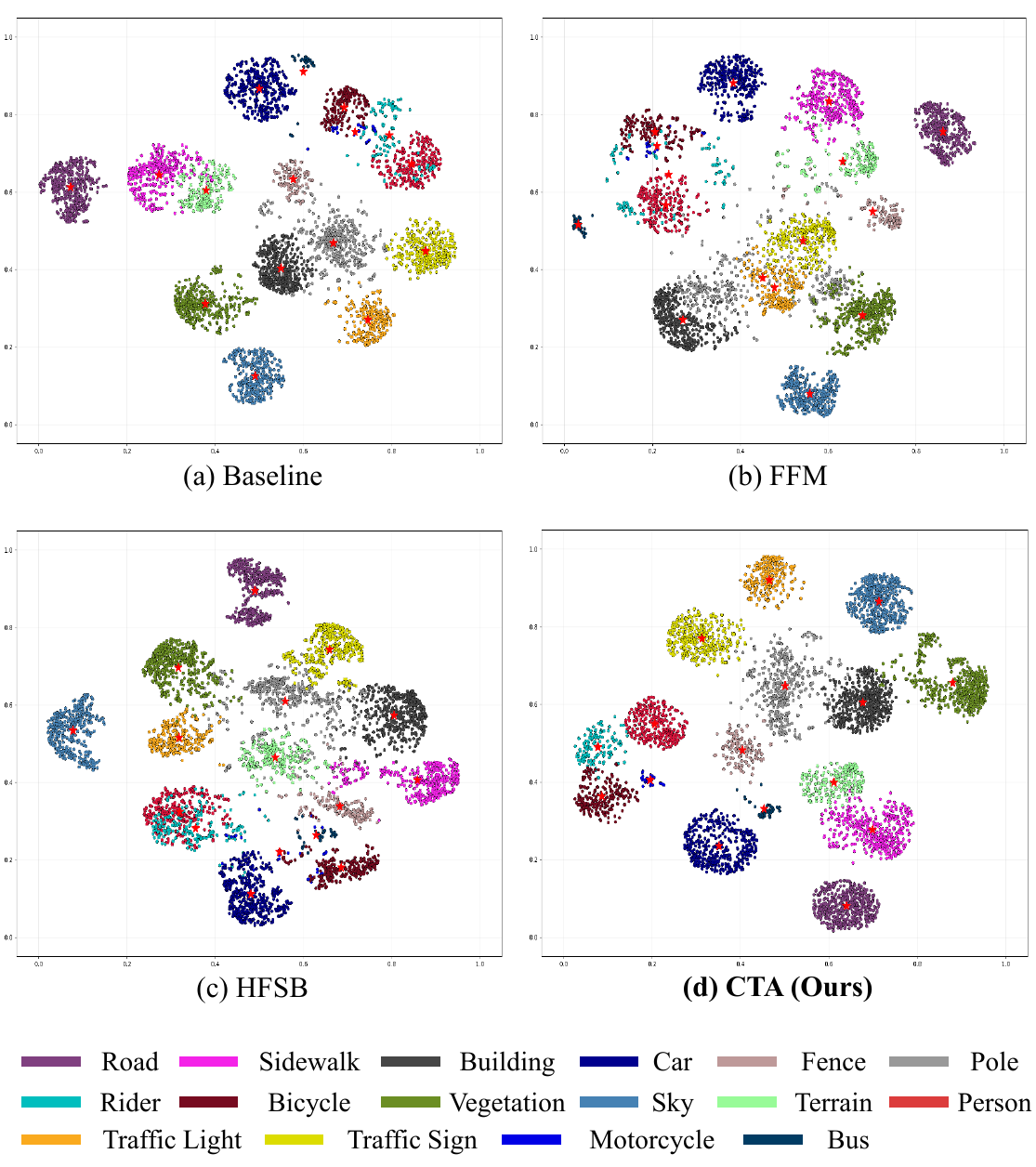}
    \caption{t-SNE visualizations of high-dimensional fused features produced by different feature fusion strategies. Red stars indicate the cluster centers corresponding to semantic categories.}
    \label{fig.tsne}
\end{figure}

\begin{table}[!t]
        \settablefont
        \centering
	\caption{
    Ablation study on the semi-supervised training strategy conducted on the KITTI 2015 dataset, where different training strategies are utilized for the geometric vision task. ``V'' , ``C'', and ``K'' are the abbreviations of the vKITTI2, Cityscapes, and KITTI 2015 datasets, respectively. 
	}
\label{tab.semi_supervised}
	\setlength{\tabcolsep}{1.5mm}
    \begin{tabular}{L{2.65cm}|C{1.15cm}C{1.15cm}|C{1.4cm} C{0.9cm}}
    \toprule
    {\multirow{2}{*}{Methods}} &  
    \multicolumn{2}{c}{Semantic Segmentation} & 
    \multicolumn{2}{c}{Stereo Matching} \\
    \cline{2-5}
    & mIoU (\%) $\uparrow$ & mFSc (\%) $\uparrow$ & EPE (pixels) $\downarrow$ & D1 (\%) $\downarrow$ \\
		\hline
            \hline
        Unsupervised (V\sra C\sra K) &68.49  &77.18 &1.58 &6.49    \\
        Semi-Supervised (V\sra K)  &54.35  &61.17  &1.09 &4.02  \\
        Semi-Supervised (V\sra C\sra K) &\textbf{72.03}  &\textbf{82.34} &\textbf{1.00} &\textbf{3.84}    \\
    	\bottomrule
		\end{tabular}
        \vspace{-1.5em}
\end{table}

\begin{table}[!t]
        \settablefont
	\centering
	\caption{
    Ablation study on different versions of the ConvNeXt backbone conducted on the Cityscapes and KITTI 2015 datasets. Inference speed is measured on an NVIDIA GeForce RTX 4090 GPU (resolution: $640\times320$ pixels).  
    }
\label{tab.backbone}
    \settablefont
	\setlength{\tabcolsep}{1.5mm}
	\begin{tabular}
	    {L{1.0cm}| C{1.2cm}  C{1.4cm} | C{1.2cm} | C{1.3cm} | C{0.9cm}}
		\toprule
		   \multicolumn{1}{l|}{\multirow{2}{*}{Backbone}} & \multicolumn{2}{c|}{KITTI 2015} & {Cityscapes} & \multirow{2}{*}{Params (M) $\downarrow$} &\multirow{2}{*}{FPS $\uparrow$} \\
           \cline{2-4}
           &mIoU (\%) $\uparrow$ & EPE (pixels)$\downarrow$ &mIoU (\%) $\uparrow$ & & \\ \hline
            {Tiny}  &70.86 &1.10 &79.49 &\textbf{67.77} &\textbf{22}\\
            {Small}  &70.81 &1.05  &79.89 &89.40 &21\\
            {Base}  &72.03 &1.00 &80.05 &137.97 &20\\
            {Large}  &\textbf{72.36} &\textbf{0.96} &\textbf{82.13} &275.89 &18\\
    	\bottomrule
		\end{tabular}
         \vspace{-1.5em}
\end{table}

\begin{table}[!t]
        \settablefont
	\centering
	\caption{
        Ablation study on the versatility and compatibility of TwInS, evaluated through its integration with other SoTA feature-fusion networks and joint learning approaches, conducted on the Cityscapes and KITTI 2015 datasets.
 	}
\label{tab.universality}
	\setlength{\tabcolsep}{1.5mm}
	\begin{tabular}
	    {L{1.8cm}| C{1.5cm}|  C{1.25cm} | C{1.25cm} | C{1.2cm}}
		\toprule
		   \multicolumn{1}{l|}{\multirow{2}{*}{Networks}} &\multirow{2}{*}{Type} & \multicolumn{1}{c}{Cityscapes} & \multicolumn{1}{c|}{KITTI 2015} & \multirow{2}{*}{Params (M) $\downarrow$} \\
           \cline{3-4}
           & &mIoU (\%) $\uparrow$ &mIoU (\%) $\uparrow$ & \\ \hline \hline
            {RTFNet}~\cite{RTFnet} & &57.89  &34.47 &254.15\\
            {\textbf{+Ours}} &{Feature-Fusion} &\textbf{63.00} &\textbf{39.10} &\textbf{199.32}\\
            \cline{1-1} \cline{3-5}
            {SNE-RoadSeg}~\cite{sneroadseg} &{Networks} &61.75 &43.21 &201.32 \\
            {\textbf{+Ours}} & &\textbf{63.50}  &\textbf{45.88} &\textbf{168.50} \\
           
           \hline
            {S$^3$M-Net}~\cite{s3mnet}& &58.29 &41.54 &346.70\\
            {\textbf{+Ours}} &{Joint Learning} &\textbf{63.91} &\textbf{45.42} &\textbf{242.68} \\
            \cline{1-1} \cline{3-5}
            {TiCoSS}~\cite{ticoss}&{Approaches} &63.57 &47.66 &385.05 \\
           {\textbf{+Ours}} & &\textbf{67.42} &\textbf{52.19} &\textbf{243.02} \\
    	\bottomrule
		\end{tabular}
        \vspace{-1.5em}
\end{table}

\subsection{Ablation Studies}
\label{sec.exp_ablation}
This subsection presents extensive ablation studies conducted to validate the effectiveness of the core components within the TwInS framework, where semantic segmentation and stereo matching are selected as the representative scene parsing and geometric vision tasks, respectively. 

\subsubsection{Context-Enriched Iterative Refinement}
Table~\ref{tab.context_enriched} presents an ablation study conducted to validate the rationality and effectiveness of incorporating contextual features from the scene parsing stream into the geometric vision stream to (i) construct the correlation pyramid, (ii) provide context-enriched features, and (iii) initialize multi-level hidden states. As reported, sequentially integrating these components yields consistent performance gains over the baseline that utilizes multiple independent encoders for separate feature extraction. The best performance is achieved when all three components are jointly used, resulting in a reduction in EPE by $11.50\%$ compared to the baseline. These results demonstrate that the contextual cues learned from the scene parsing task can effectively guide the iterative refinement process, ultimately improving the performance of the geometric vision task.

\subsubsection{Cross-Task Adapter} 
Table~\ref{tab.cross_task_adapter} presents an ablation study conducted to validate the effectiveness of the proposed CTA, compared with the element-wise addition-based feature fusion module (FFM) from S$^3$M-Net~\cite{s3mnet} and the self-attention-based heterogeneous feature synergy block (HFSB) from RoadFormer~\cite{roadformer}. The baseline model only utilizes the contextual features for scene parsing, without incorporating cross-view geometric cues from the geometric vision stream. 

These results suggest that the proposed CTA significantly outperforms the baseline, improving the mIoU by up to $5.63\%$. This performance gain underscores the effectiveness of utilizing iteratively refined hidden states from the geometric vision task to enhance scene parsing via CTA. 
In contrast, HFSB underperforms the baseline, reducing the mIoU by more than $2.67\%$. This performance degradation can be primarily attributed to HFSB's architectural design, which is tailored for fusing heterogeneous features extracted by duplex encoders and is thus ill-suited for task-specific feature fusion. 
The t-SNE~\cite{tsne_2008_jmlr} visualization results shown in Fig.~\ref{fig.tsne} further support this claim.
Unlike CTA, HFSB fails to generate compact and well-separated clusters in the high-dimensional feature space, even performing worse than the baseline. This limitation is particularly evident in semantically ambiguous categories such as rider, person, and bicycle, where the inappropriate infusion of geometric features disrupts the original contextual features. In contrast, CTA markedly improves both intra-class compactness and inter-class separability of the fused features, thereby leading to superior scene parsing performance.

\subsubsection{Semi-Supervised Training Strategy}
To demonstrate the necessity and effectiveness of the proposed semi-supervised training strategy, Table~\ref{tab.semi_supervised} presents quantitative comparisons of three training paradigms for the geometric vision task, evaluated on the KITTI 2015 dataset: (i) unsupervised training, sequentially conducted on the VKITTI2, Cityscapes, and KITTI 2015 datasets, (ii) direct semi-supervised fine-tuning on the KITTI 2015 dataset following supervised pre-training on the VKITTI2 dataset, and (iii) semi-supervised training, sequentially conducted on the Cityscapes and KITTI 2015 datasets, after supervised pre-training on the VKITTI2 dataset.

The comparison between paradigms (i) and (iii) underscores the superiority of the proposed semi-supervised training strategy, which effectively extracts valuable geometric cues from additional unlabeled data. In contrast, when fully unsupervised training is performed on newly introduced unlabeled data, the performance of the geometric vision task deteriorates. This degradation results in unreliable geometric features, which in turn adversely affect the scene parsing stream, highlighting the importance of guided semi-supervised learning in maintaining overall system robustness and accuracy. On the other hand, the comparison between paradigms (ii) and (iii) demonstrates the self-evolving capability of the proposed semi-supervised strategy, which effectively unleashes the potential of large-scale multi-view data and further improves the performance of the overall joint learning framework. An additional experiment regarding the selection of the confidence percentile $\alpha$ is provided in the supplement. 

\subsubsection{Efficiency and Accuracy Trade-off}
An ablation study on different backbone versions is also conducted to evaluate the balance between the accuracy and computational efficiency of the TwInS framework. As expected,  TwInS yields the best performance in both scene parsing and geometric vision tasks when equipped with a large backbone, as shown in Table~\ref{tab.backbone}. 
Interestingly, TwInS equipped with a tiny backbone reduces model parameters by over $75.43\%$ and increases inference speed by $21.75\%$, compared to that with the large backbone, without compromising accuracy. 

\subsubsection{Versatility and Compatibility}
Finally, the ablation study presented in Table~\ref{tab.universality} demonstrates the strong versatility and compatibility of the TwInS framework, which consistently improves the performance of various SoTA feature-fusion networks and joint learning approaches. 
Notably, integrating TiCoSS within the TwInS framework improves the mIoU by up to $9.50\%$, while simultaneously reducing the total model parameters by $36.88\%$. These results underscore that TwInS can serve as a universal framework to further improve semantic segmentation performance, particularly for approaches that rely on heavy duplex encoders. 

\begin{figure}[!t]
	\centering
	\includegraphics[width=0.49\textwidth]{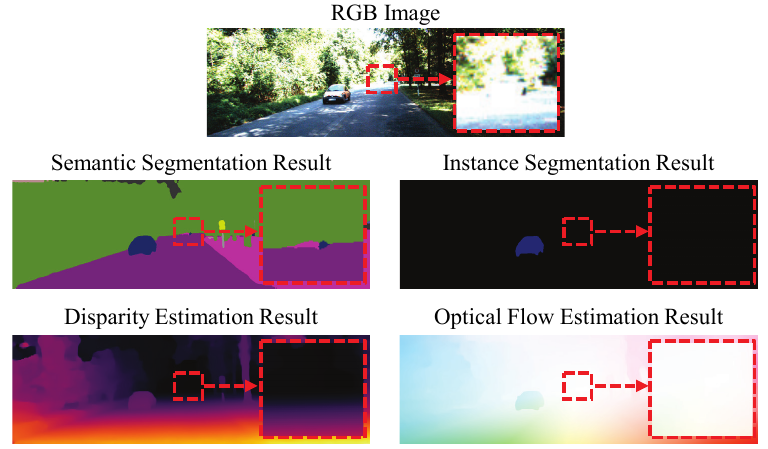}
	\caption{
    A failure case of TwInS, where the vehicle in the overexposed region is misidentified as a result of unreliable correspondence matching.
    }
    \vspace{-1.5em}
	\label{fig.failure_case}
\end{figure}

\subsection{Limitations}
Despite achieving compelling results, TwInS still exhibits two main limitations. First, the geometric vision task plays a dominant role in the joint learning framework. As shown in Fig.~\ref{fig.failure_case}, under poor illumination conditions, the geometric vision stream may fail to produce reliable disparity or optical flow estimations, which in turn degrades the performance of the scene parsing task. This observation underscores the need for more robust task interaction mechanisms that can mitigate the adverse influence of one task’s failure on the other. Second, although the proposed semi-supervised training strategy effectively eliminates the need for correspondence annotations in newly introduced datasets, it still relies on segmentation annotations. Extending the TwInS framework to open-vocabulary segmentation paradigms therefore emerges as a promising and necessary direction for future research, aiming to further reduce annotation dependency and enhance the framework’s adaptability to diverse, large-scale datasets.

\section{Conclusion and Future Work}
This article introduced TwInS, a novel general-purpose joint learning framework developed to simultaneously tackle scene parsing and geometric vision tasks through two bidirectionally interactive streams. The scene parsing stream performs semantic or instance segmentation, where multi-level contextual features were utilized for context-enriched iterative refinement in the geometric vision stream. Concurrently, the geometric vision stream estimates correspondences in a coarse-to-fine manner, with its decoded geometric features selectively fused with contextual features via a novel cross-task adapter, thereby enabling more comprehensive driving scene parsing. Moreover, a semi-supervised training strategy based on uncertainty-aware pseudo correspondence generation was developed for geometric vision tasks, facilitating the effective incorporation of large-scale multi-view data. The effectiveness of each proposed component as well as the superior overall performance of TwInS were validated through extensive experiments conducted on both synthetic and real-world datasets. 

Future work will focus on developing effective cross-task mutual supervision mechanisms to establish a fully self-evolving joint learning framework, with the ultimate goal of minimizing or even eliminating the reliance on human annotations, thereby paving the way for more autonomous and advanced visual perception systems.
\label{sec.conclusion}
\bibliographystyle{IEEEtran}
\bibliography{refs}

@String(CVPR= {IEEE Conf. Comput. Vis. Pattern Recog.})

@String(ICCV= {Int. Conf. Comput. Vis.})

@String(ECCV= {Eur. Conf. Comput. Vis.})

@String(ICLR = {Int. Conf. Learn. Represent.})

@String(AAAI = {AAAI})

@incollection{twostream1,
  author    = {Ungerleider, Leslie G.},
  title     = {{Two Cortical Visual Systems}},
  booktitle = {Analysis of Visual Behavior},
  chapter   = {18},
  pages     = {549--586},
  year      = {1982},
  publisher = {MIT Press}
}

@article{bisenetv2,
  title={{BiSeNet V2: Bilateral Network with Guided Aggregation for Real-Time Semantic Segmentation}},
  author={Yu, Changqian and others},
  journal={International Journal of Computer Vision},
  volume={129},
  pages={3051--3068},
  year={2021},
  publisher={Springer}
}

@inproceedings{segmenter,
  title={{Segmenter: Transformer for Semantic Segmentation}},
  author={Strudel, Robin and others},
  booktitle={Proceedings of the IEEE/CVF International Conference on Computer Vision (ICCV)},
  pages={7262--7272},
  year={2021}
}

@article{segformer,
  title={{SegFormer: Simple and Efficient Design for Semantic Segmentation with Transformers}},
  author={Xie, Enze and others},
  journal={Proceedings of the Advances in Neural Information Processing Systems (NeurIPS)},
  volume = {34},
  pages={12077--12090},
  year={2021}
}

@article{knet,
  title={{K-Net: Towards Unified Image Segmentation}},
  author={Zhang, Wenwei and others},
  journal={Proceedings of the Advances in Neural Information Processing Systems (NeurIPS)},
  pages={10326--10338},
  volume = {34},
  year={2021}
}

@inproceedings{mask2former,
  title={{Masked-Attention Mask Transformer for Universal Image Segmentation}},
  author={Cheng, Bowen and others},
  booktitle={Proceedings of the IEEE/CVF Conference on Computer Vision and Pattern Recognition (CVPR)},
  pages={1290--1299},
  year={2022}
}

@inproceedings{dformer,
  title={{DFormer: Rethinking RGBD Representation Learning for Semantic Segmentation}},
  author={Yin, Bowen and others},
  booktitle={International Conference on Learning Representations (ICLR)},
  year={2024}
}

@inproceedings{dformerv2,
  title={{DFormerv2: Geometry Self-Attention for RGBD Semantic Segmentation}},
  author={Yin, Bo-Wen and others},
  booktitle={Proceedings of the IEEE/CVF Conference on Computer Vision and Pattern Recognition (CVPR)},
  pages={19345--19355},
  year={2025}
}

@article{mtl_review1_tpami,
  title={{Multi-Task Learning for Dense Prediction Tasks: A Survey}},
  author={Vandenhende, Simon and others},
  journal={IEEE Transactions on Pattern Analysis and Machine Intelligence},
  volume={44},
  number={7},
  pages={3614--3633},
  year={2021},
  publisher={IEEE}
}

@inproceedings{linear_attention,
  title={{Transformers are RNNs: Fast Autoregressive Transformers with Linear Attention}},
  author={Katharopoulos, Angelos and others},
  booktitle={Proceedings of the International Conference on Machine Learning (ICML)},
  pages={5156--5165},
  year={2020},
}

@InProceedings{cst_stereo,
    author    = {Zhou, Jingyi and others},
    title     = {{Consistency-aware Self-Training for Iterative-based Stereo Matching}},
    booktitle = {Proceedings of the IEEE/CVF Conference on Computer Vision and Pattern Recognition (CVPR)},
    year      = {2025},
    pages     = {16641-16650}
}

@inproceedings{semi_stereo,
  title={{Semi-Stereo: A Universal Stereo Matching Framework for Imperfect Data via Semi-supervised Learning}},
  author={Yue, Xin and others},
  booktitle={Proceedings of the IEEE/CVF Conference on Computer Vision and Pattern Recognition (CVPR) Workshops},
  pages={646--655},
  year={2024}
}

@inproceedings{dualnet_wang_aaai25,
  title={{DualNet: Robust Self-Supervised Stereo Matching with Pseudo-Label Supervision}},
  author={Wang, Yun and others},
  booktitle={Proceedings of the AAAI Conference on Artificial Intelligence (AAAI)},
  pages={8178--8186},
  year={2025}
}

@misc{dinov2,
  title={{DINOv2: Learning Robust Visual Features without Supervision}},
  author={Oquab, Maxime and others},
  archivePrefix={arXiv},
  note={{\it {arXiv:2304.07193}}},
  year={2023}
}

@misc{vkitti2,
  title={{Virtual KITTI 2}},
  author={Cabon, Yohann and others},
  year={2020},
  note={{\it {arXiv:2304.07193}}},
}

@article{RTFnet,
  title={{RTFNet: RGB-Thermal Fusion Network for Semantic Segmentation of Urban Scenes}},
  author={Sun, Yuxiang and others},
  journal={IEEE Robotics and Automation Letters},
  volume={4},
  number={3},
  pages={2576--2583},
  year={2019},
  publisher={IEEE}
}

@inproceedings{sednet_chen_cvpr23,
  title={{Learning the Distribution of Errors in Stereo Matching for Joint Disparity and Uncertainty Estimation}},
  author={Chen, Liyan and others},
  booktitle={Proceedings of the IEEE/CVF Conference on Computer Vision and Pattern Recognition (CVPR)},
  pages={17235--17244},
  year={2023}
}

@article{kendall_2017_uncertainties,
  title={{What Uncertainties Do We Need in Bayesian Deep Learning for Computer Vision?}},
  author={Kendall, Alex and Gal, Yarin},
  journal={Proceedings of the Advances in Neural Information Processing Systems (NeurIPS)},
  volume={30},
  year={2017}
}

@inproceedings{Rad2021ICML,
  title={{Learning Transferable Visual Models From Natural Language Supervision}},
  author={Radford, Alec and others},
  booktitle={Proceedings of the 38th International Conference on Machine Learning (ICML)},
  pages={8748--8763},
  year={2021},
}

@inproceedings{convnext,
  title={{A ConvNet for the 2020s}},
  author={Liu, Zhuang and others},
  booktitle={Proceedings of the IEEE/CVF conference on Computer Vision and Pattern Recognition (CVPR)},
  pages={11976--11986},
  year={2022}
}

@article{tsne_2008_jmlr,
  title={{Visualizing Data using t-SNE}},
  author={Maaten, Laurens van der and Hinton, Geoffrey},
  journal={Journal of Machine Learning Research},
  volume={9},
  pages={2579--2605},
  year={2008}
}

@inproceedings{flow2stereo,
  title={{Flow2Stereo: Effective Self-Supervised Learning of Optical Flow and Stereo Matching}},
  author={{P. Liu} and others},
  booktitle={Proceedings of the IEEE/CVF conference on Computer Vision and Pattern Recognition (CVPR)},
  pages={6648--6657},
  year={2020}
}

@inproceedings{selflow,
    title = {{SelFlow: Self-Supervised Learning of Optical Flow}},
    author = {Liu, Pengpeng and others},
    booktitle = {Proceedings of the IEEE/CVF Conference on Computer Vision and Pattern Recognition (CVPR)},
    pages={4571--4580},
    year = {2019}
}

@inproceedings{CRD-Fusion,
  title={{Occlusion-Aware Self-Supervised Stereo Matching with Confidence Guided Raw Disparity Fusion}},
  author={Fan, Xiule and others},
  booktitle={Proceedings of the Conference on Robots and Vision (CRV)},
  pages={132--139},
  year={2022},
  organization={IEEE}
}

@inproceedings{es3net,
  title={{ES3Net: Accurate and Efficient Edge-Based Self-Supervised Stereo Matching Network}},
  author={Fang, I and others},
  booktitle={Proceedings of the IEEE/CVF Conference on Computer Vision and Pattern Recognition (CVPR)},
  pages={4472--4481},
  year={2023}
}

@article{LiDAR-camera,
  title={{Automatic Targetless LiDAR-Camera Calibration: A Survey}},
  author={Li, Xingchen and others},
  journal={Artificial Intelligence Review},
  volume={56},
  number={9},
  pages={9949--9987},
  year={2023},
  publisher={Springer}
}

@inproceedings{upflow,
  title={{UPFlow: Upsampling Pyramid for Unsupervised Optical Flow Learning}},
  author={Luo, Kunming and others},
  booktitle={Proceedings of the IEEE/CVF Conference on Computer Vision and Pattern Recognition (CVPR)},
  pages={1045--1054},
  year={2021}
}

@inproceedings{semstereo_chen_aaai2025,
  title={{SemStereo: Semantic-Constrained Stereo Matching Network for Remote Sensing}},
  author={Chen, Chen and others},
  booktitle={Proceedings of the AAAI Conference on Artificial Intelligence (AAAI)},
  pages={15758--15766},
  year={2025}
}

@inproceedings{SGDepth,
  title={{Self-supervised Monocular Depth Estimation: Solving the Dynamic Object Problem by Semantic Guidance}},
  author={Klingner, Marvin and others},
  booktitle={Proceedings of the European Conference on Computer Vision (ECCV)},
  pages={582--600},
  year={2020},
}

@ARTICLE{vitas,
  author={Liu, Chuang-Wei and others},
  journal={IEEE Transactions on Intelligent Vehicles}, 
  title={{Playing to Vision Foundation Model's Strengths in Stereo Matching}}, 
  year={2024},
  volume={},
  number={},
  pages={1-12},
  note={{DOI}: 10.1109/TIV.2024.3467287}}

@inproceedings{psmnet,
  title={{Pyramid Stereo Matching Network}},
  author={Chang, Jia-Ren and Chen, Yong-Sheng},
  booktitle={Proceedings of the IEEE/CVF Conference on Computer Vision and Pattern Recognition (CVPR)},
  pages={5410--5418},
  year={2018}
}

@InProceedings{dcflow,
author = {Xu, Jia and others},
title = {{Accurate Optical Flow via Direct Cost Volume Processing}},
booktitle = {Proceedings of the IEEE Conference on Computer Vision and Pattern Recognition (CVPR)},
pages={1289--1297},
year = {2017}
}

@inproceedings{raftstereo,
  title={{RAFT-Stereo: Multilevel Recurrent Field Transforms for Stereo Matching}},
  author={Lipson, Lahav and others},
  booktitle={International Conference on 3D Vision (3DV)},
  pages={218--227},
  year={2021},
  organization={IEEE}
}

@inproceedings{raft,
  title={{RAFT: Recurrent All-Pairs Field Transforms for Optical Flow}},
  author={Teed, Zachary and Deng, Jia},
  booktitle={Proceedings of the European Conference on Computer Vision (ECCV)},
  pages={402--419},
  year={2020},
}

@inproceedings{depthanything,
  title={{Depth Anything: Unleashing the Power of Large-Scale Unlabeled Data}},
  author={Yang, Lihe and others},
  booktitle={Proceedings of the IEEE/CVF conference on Computer Vision and Pattern Recognition (CVPR)},
  pages={10371--10381},
  year={2024}
}

@inproceedings{vit_comer,
  title={{ViT-CoMer: Vision Transformer with Convolutional Multi-scale Feature Interaction for Dense Predictions}},
  author={Xia, Chunlong and others},
  booktitle={Proceedings of the IEEE/CVF conference on Computer Vision and Pattern Recognition (CVPR)},
  pages={5493--5502},
  year={2024}
}

@ARTICLE{cmxnet,
  author={Zhang, Jiaming and others},
  journal={IEEE Transactions on Intelligent Transportation Systems}, 
  title={{CMX: Cross-Modal Fusion for RGB-X Semantic Segmentation With Transformers}}, 
  year={2023},
  volume={24},
  number={12},
  pages={14679-14694},
  doi={10.1109/TITS.2023.3300537}}

@ARTICLE{guo2024lix,
  author={Guo, Sicen and others},
  journal={IEEE Transactions on Image Processing}, 
  title={{LIX: Implicitly Infusing Spatial Geometric Prior Knowledge Into Visual Semantic Segmentation for Autonomous Driving}}, 
  year={2025},
  volume={34},
  number={},
  pages={7250-7263},
  doi={10.1109/TIP.2025.3618378}}

@inproceedings{segstereo,
  title={{SegStereo: Exploiting Semantic Information for Disparity Estimation}},
  author={Yang, Guorun and others},
  booktitle={Proceedings of the European Conference on Computer Vision (ECCV)},
  pages={636--651},
  year={2018}
}

@article{DispSegNet,
  title={{DispSegNet: Leveraging Semantics for End-to-end Learning of Disparity Estimation from Stereo Imagery}},
  author={Zhang, Junming and others},
  journal={IEEE Robotics and Automation Letters},
  year={2018},
  volume={4},
  pages={1162-1169},
}

@INPROCEEDINGS{dsnet,
  author={Zhan, Wujing and others},
  booktitle={Proceedings of the International Conference on Robotics and Automation (ICRA)}, 
  title={{DSNet: Joint Learning for Scene Segmentation and Disparity Estimation}}, 
  year={2019},
  volume={},
  number={},
  pages={2946-2952},
  doi={10.1109/ICRA.2019.8793573}}

@inproceedings{kitti2015,
  title={{Object Scene Flow for Autonomous Vehicles}},
  author={Menze, Moritz and Geiger, Andreas},
  booktitle={Proceedings of the IEEE/CVF Conference on Computer Vision and Pattern Recognition (CVPR)},
  pages={3061--3070},
  year={2015}
}

@inproceedings{igevstereo,
  title={{Iterative Geometry Encoding Volume for Stereo Matching}},
  author={Xu, Gangwei and others},
  booktitle={Proceedings of the IEEE/CVF Conference on Computer Vision and Pattern Recognition (CVPR)},
  pages={21919--21928},
  year={2023}
}

@inproceedings{flowformer,
  title={{Flowformer: A Transformer Architecture for Optical Flow}},
  author={Huang, Zhaoyang and others},
  booktitle={European Conference on Computer Vision (ECCV)},
  pages={668--685},
  year={2022},
}

@inproceedings{sneroadseg,
  title={{SNE-RoadSeg: Incorporating Surface Normal Information into Semantic Segmentation for Accurate Freespace Detection}},
  author={Fan, Rui and others},
  booktitle={Proceedings of the European Conference on Computer Vision (ECCV)},
  pages={340--356},
  year={2020},
}

@inproceedings{rts2net,
  title={{Real-time Semantic Stereo Matching}},
  author={Dovesi, Pier Luigi and others},
  booktitle={Proceedings of the International Conference on Robotics and Automation (ICRA)},
  pages={10780--10787},
  year={2020},
}

@inproceedings{cityscapes,
  title={{The Cityscapes Dataset for Semantic Urban Scene Understanding}},
  author={Cordts, Marius and others},
  booktitle={Proceedings of the IEEE/CVF Conference on Computer Vision and Pattern Recognition (CVPR)},
  pages={3213--3223},
  year={2016}
}

@ARTICLE{roadformer,
  author={Li, Jiahang and others},
  journal={IEEE Transactions on Intelligent Vehicles}, 
  title={{RoadFormer: Duplex Transformer for RGB-Normal Semantic Road Scene Parsing}}, 
  year={2024},
  volume={9},
  number={7},
  pages={5163-5172},
  doi={10.1109/TIV.2024.3388726}}

@ARTICLE{roadformer+,
  author={Huang, Jianxin and others},
  journal={IEEE Transactions on Intelligent Vehicles}, 
  title={{RoadFormer+: Delivering RGB-X Scene Parsing through Scale-Aware Information Decoupling and Advanced Heterogeneous Feature Fusion}}, 
  year={2025},
  volume={10},
  number={5},
  pages={3156-3165},
  doi={10.1109/TIV.2024.3448251}}

@ARTICLE{ticoss,
  author={Tang, Guanfeng and others},
  journal={IEEE Transactions on Automation Science and Engineering}, 
  title={{TiCoSS: Tightening the Coupling Between Semantic Segmentation and Stereo Matching Within a Joint Learning Framework}}, 
  year={2025},
  volume={22},
  number={},
  pages={18646-18658},
  doi={10.1109/TASE.2025.3586286}}

@inproceedings{mask_rcnn,
  title={{Mask R-CNN}},
  author={He, Kaiming and others},
  booktitle={Proceedings of the IEEE International Conference on Computer Vision (ICCV)},
  pages={2961--2969},
  year={2017}}

@article{maskformer,
  title={{Per-Pixel Classification is Not All You Need for Semantic Segmentation}},
  author={Cheng, Bowen and others},
  journal={Proceedings of the Advances in Neural Information Processing Systems (NeurIPS)},
  volume = {34},
  pages={17864--17875},
  year={2021}
}

@inproceedings{SparseInst,
  title={{Sparse Instance Activation for Real-Time Instance Segmentation}},
  author={Cheng, Tianheng and others},
  booktitle={Proceedings of the IEEE/CVF conference on Computer Vision and Pattern Recognition (CVPR)},
  pages={4433--4442},
  year={2022}
}

@article{calibnet,
  title={{CalibNet: Dual-Branch Cross-Modal Calibration for RGB-D Salient Instance Segmentation}},
  author={Pei, Jialun and others},
  journal={IEEE Transactions on Image Processing},
  year={2024},
  volume={33},
  number={},
  pages={4348-4362},
  publisher={IEEE}
}

@inproceedings{semarflow,
  title={{SemARFlow: Injecting Semantics into Unsupervised Optical Flow Estimation for Autonomous Driving}},
  author={Yuan, Shuai and others},
  booktitle={Proceedings of the IEEE/CVF International Conference on Computer Vision (CVPR)},
  pages={9566--9577},
  year={2023}
}

@ARTICLE{rgbd_sensors_tip,
  author={Liu, Xianming and others},
  journal={IEEE Transactions on Image Processing}, 
  title={{Depth Restoration From RGB-D Data via Joint Adaptive Regularization and Thresholding on Manifolds}}, 
  year={2019},
  volume={28},
  number={3},
  pages={1068-1079},
  doi={10.1109/TIP.2018.2872175}}

@article{correspondence_tip,
  title={{Detail Preserving Coarse-to-Fine Matching for Stereo Matching and Optical Flow}},
  author={Deng, Yong and others},
  journal={IEEE Transactions on Image Processing},
  volume={30},
  pages={5835--5847},
  year={2021},
  publisher={IEEE}
}

@article{rose_wang_tcsvt25,
  title={{RoSe: Robust Self-supervised Stereo Matching under Adverse Weather Conditions}},
  author={Wang, Yun and others},
  journal={IEEE Transactions on Circuits and Systems for Video Technology},
  year={2025},
  publisher={IEEE},
  note={{DOI}= 10.1109/TCSVT.2025.3622186}
}

@INPROCEEDINGS{sgroadseg,
  author={Wu, Zhiyuan and others},
  booktitle={Proceedings of the International Conference on Robotics and Automation (ICRA)}, 
  title={{SG-RoadSeg: End-to-End Collision-Free Space Detection Sharing Encoder Representations Jointly Learned via Unsupervised Deep Stereo}}, 
  year={2024},
  volume={},
  number={},
  pages={11063-11069},
  doi={10.1109/ICRA57147.2024.10611191}}

@ARTICLE{sgroadseg+,
  author={Lee, Ming-Ju and others},
  journal={IEEE Transactions on Instrumentation and Measurement}, 
  title={{SG-RoadSeg+: End-to-End Freespace Detection Upgraded at Data, Feature, and Loss Levels}}, 
  year={2025},
  volume={74},
  number={},
  pages={1-9},
  doi={10.1109/TIM.2025.3579733}}

@inproceedings{ARflow,
  title={{Learning by Analogy: Reliable Supervision From Transformations for Unsupervised Optical Flow Estimation}},
  author={Liu, Liang and others},
  booktitle={Proceedings of the IEEE/CVF Conference on Computer Vision and Pattern Recognition (CVPR)},
  pages={6489--6498},
  year={2020}
}

@article{s3mnet,
  title={{S$^3$M-Net: Joint Learning of Semantic Segmentation and Stereo Matching for Autonomous Driving}},
  author={{Z. Wu} and others},
  journal={IEEE Transactions on Intelligent Vehicles},
  volume={9},
  number={2},
  pages={3940--3951},
  year={2024},
  publisher={IEEE}
}
\nolinenumbers
\end{document}